\documentclass[runningheads]{llncs}

\usepackage{eccv}

\usepackage[table]{xcolor}
\newcommand{\first}[1]{\cellcolor{red!25}#1}
\newcommand{\second}[1]{\cellcolor{orange!30}#1}
\newcommand{\third}[1]{\cellcolor{yellow!30}#1}

\usepackage[normalem]{ulem}
\usepackage{siunitx}
\DeclareSIUnit\pixel{px}

\usepackage{amsmath}

\DeclareMathOperator*{\argmin}{arg\,min}

\usepackage{eccvabbrv}

\usepackage{graphicx}
\usepackage{booktabs}
\usepackage{array}
\usepackage{placeins}

\usepackage[accsupp]{axessibility}  %

\usepackage{hyperref}

\usepackage{orcidlink}

\usepackage{marvosym}

\begin{document}

\title{Gaussians on Fire: High-Frequency Reconstruction of Flames}

\author{Jakob Nazarenus\inst{1\text{(\Letter)}}\orcidlink{0000-0002-6800-2462} \and
Dominik Michels\inst{2}\orcidlink{0000-0002-1621-325X} \and
Wojtek Palubicki\inst{3}\orcidlink{0000-0002-2374-346X} \and Simin Kou\inst{4}\orcidlink{0000-0002-7222-2214} \and Fang-Lue Zhang\inst{4}\orcidlink{0000-0002-8728-8726} \and
Sören Pirk\inst{1}\orcidlink{0000-0003-1937-9797} \and Reinhard Koch\inst{1}\orcidlink{0000-0003-4398-1569}}

\authorrunning{J.~Nazarenus et al.}

\institute{Kiel University, Kiel, Germany\\
\email{\{jna,sp,rk\}@informatik.uni-kiel.de} \and
KAUST, CEMSE, Thuwal, Saudi Arabia\\
\email{dominik.michels@kaust.edu.sa} \and 
Adam Mickiewicz University, Poznań, Poland\\
\email{Wojciech.Palubicki@amu.edu.pl} \and
Victoria University of Wellington, Wellington, New Zealand\\
\email{\{simin.kou,fanglue.zhang\}@vuw.ac.nz}}

\maketitle
\makeatletter\begingroup\renewcommand\thefootnote{}\long\def\@makefntext#1{\noindent#1}\footnotetext{%
This arXiv version has corrected results for our method (see footnote in Section~\ref{sec:results}).\\
\copyright~The Author(s) 2026, \href{https://creativecommons.org/licenses/by/4.0/}{CC BY 4.0}\\
P. Favaro et al. (Eds.): ECCV 2026, LNCS 17068, pp. 454--476, 2026.\\
\url{https://doi.org/10.1007/978-3-032-37152-2_25}}\endgroup\makeatother
\setcounter{footnote}{0}

\begin{abstract}
  We propose a method to reconstruct dynamic fire in 3D from a limited set of camera views with a Gaussian-based spatiotemporal representation. Capturing and reconstructing fire and its dynamics is highly challenging due to its volatile nature, transparent quality, and multitude of high-frequency features. Despite these challenges, we aim to reconstruct fire from only three views, which consequently requires solving for under-constrained geometry. We solve this by separating the static background from the dynamic fire region by combining dense multi-view stereo images with monocular depth priors. The fire is initialized as a 3D flow field, obtained by fusing per-view dense optical flow projections. To capture the high-frequency features of fire, each 3D Gaussian encodes a lifetime and linear velocity to match the dense optical flow. To ensure sub-frame temporal alignment across cameras, we employ a custom hardware synchronization pattern -- allowing us to reconstruct fire with affordable commodity hardware. Our quantitative and qualitative validations across numerous reconstruction experiments demonstrate robust performance for diverse and challenging real and simulated fire scenarios.
\end{abstract}

\section{Introduction}

\begin{figure}[t]
    \centering
    \includegraphics[clip, trim=0.5cm 1cm 1cm 0.5cm, width=\linewidth]{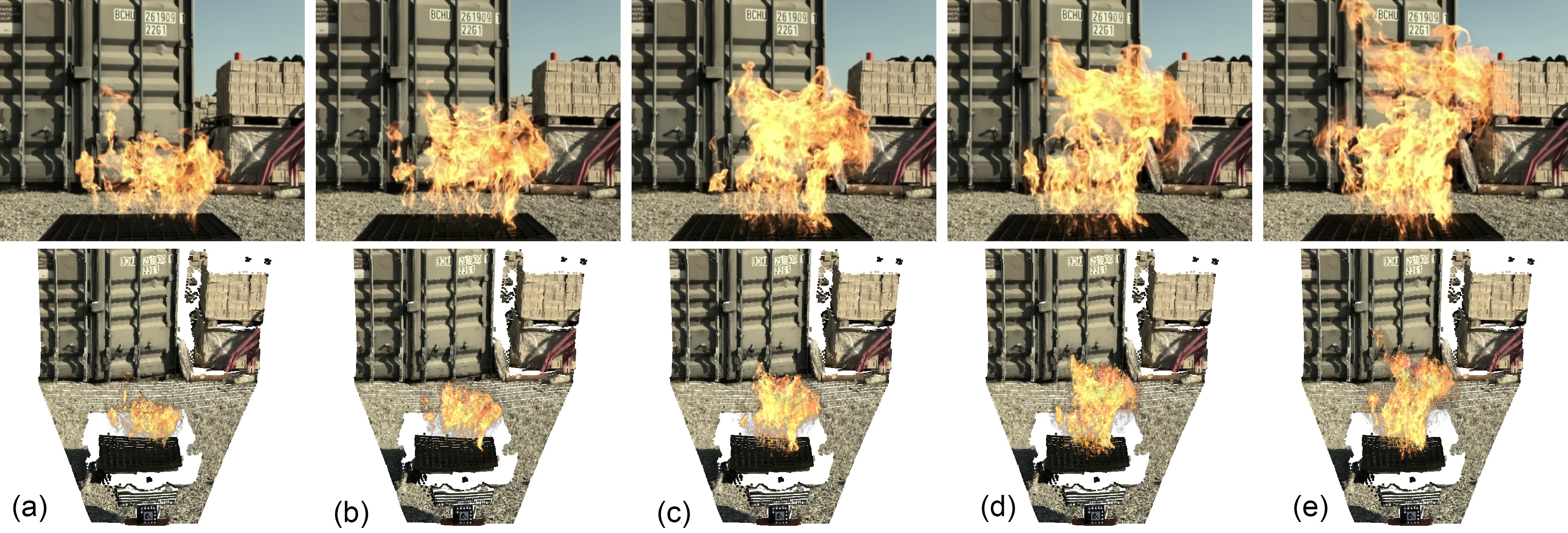}
    \caption{Temporal progression of a flame encoded as 3D Gaussian splats. The sequence shows the evolution from an initial small flame~(a, b) to a fully developed flame~(c, d, e) as the rendered 3D Gaussian splats capturing volumetric radiance and motion (top row) and the corresponding point clouds highlighting the spatial distribution and density of the Gaussian primitives over time (bottom row).}
    \vspace{-6mm}
    \label{fig:teaser}
\end{figure}

In recent years, differentiable scene representations have enabled remarkable progress in reconstructing static and dynamic 3D environments. In particular, Gaussian Splatting~\cite{Kerbl2023GaussianSplatting} approaches have gained popularity due to their good performance and efficient reconstruction run-times. However, reconstructing highly dynamic objects remains challenging due to rapid motion, appearance changes, and the lack of consistent geometry across time. 
Capturing fire, in particular, is uniquely difficult because of its non-rigid structure, high-frequency appearance, and rapid temporal variation.
Current Gaussian Splatting-based approaches fundamentally fail when applied to fire, a limitation we verify across different experiments. We compare against state-of-the-art dynamic GS methods, including 4D Gaussian Splatting \cite{Wu2024FourDGS, yang2023gs4d} and FreeTimeGS~\cite{Wang2025FreeTimeGS}, and observe that all baselines produce degenerate reconstructions for flames: geometry collapses, motion becomes inconsistent, and rendered depth fields are spatially unstable. These methods either blur away the flame structure or incorrectly attribute it to background geometry, resulting in poor reconstruction quality in dynamic regions~(\cref{fig:compositions,tab:results-real}). Across these baselines, we observe three recurring failure modes: (i) geometric absorption, where flame motion is explained by deforming the background; (ii) depth collapse, where plausible image appearance is obtained without a meaningful 3D structure; and (iii) temporal drift, where the rendered flame no longer matches the observed dynamic evolution. Our method is designed specifically to avoid these three modes. This failure stems from a shared assumption across existing GS methods -- that scene content can be represented as persistent primitives undergoing smooth, trackable motion. Fire violates this assumption entirely: it is transient, lacks consistent correspondences, and exhibits strongly view-dependent emission that cannot be explained by surface-based radiance models. Consequently, no existing Gaussian Splatting method is able to reliably reconstruct fire from real-world observations, particularly under sparse-view conditions. This exposes a critical gap in current discrete dynamic Gaussian scene representations and motivates the need for a formulation specifically designed for highly dynamic, emissive, and volumetric phenomena. The key representational change in our approach is that we do not treat flames as persistent scene elements undergoing deformation. Instead, we model fire as a population of transient volumetric carriers of emission whose motion and temporal support must be inferred jointly. This shift in the unit of representation is what makes sparse-view reconstruction of flames feasible within a Gaussian Splatting framework.

We introduce a pipeline that reconstructs high-frequency spatiotemporal flame dynamics with 3D Gaussian Splats. As it is challenging to capture fire in real-world environments, our method targets a highly constrained, sparse-view setup. Specifically, we present an approach for capturing fire with only three calibrated cameras. Our approach is based on the idea to first decouple the static scene from the flames. We then initialize Gaussian splats for fire based on a 3D flow field fused from per-view dense optical flow. Additionally, we use an explicit temporal parameterization -- each Gaussian carries a lifetime and linear velocity -- which enables motion priors (e.g., upward bias, smoothness) under sparse views. We refer to these dynamic primitives as transient flame Gaussians: short-lived volumetric carriers of emission whose finite temporal support is essential for modeling fire. This flow-initialized representation couples naturally with efficient Gaussian rasterization, yielding stable optimization and high-fidelity renderings despite the severe view sparsity.

Our method reconstructs temporally coherent 4D representations of fire that capture both its volumetric appearance and dynamic motion from sparse-view video input. Given as few as three calibrated and synchronized cameras, the pipeline jointly estimates a static background scene and a time-varying Gaussian field representing the flames. The resulting model allows photo-realistic novel view synthesis, temporal re-rendering at arbitrary time steps, and explicit analysis of motion trajectories through the per-Gaussian velocity field. In practice, this enables stable reconstruction across diverse flame types -- from small localized flames to larger outdoor fires -- without requiring dense camera arrays or simulation-based priors. Importantly, our formulation is intentionally minimal: rather than introducing a heavy physics simulator or a fully new rendering backbone, we show that rethinking decomposition, motion initialization, and temporal support is sufficient to make sparse-view fire reconstruction tractable.

A result of our method is shown in Figure \ref{fig:teaser}. In summary, our work makes the following contributions:
(1) We identify sparse-view fire reconstruction as a failure case for existing dynamic GS methods and analyze this failure in terms of their persistent-primitive and smooth-motion assumptions;
(2) We introduce a transient Gaussian reconstruction formulation that decouples static scene geometry from short-lived emissive flame structures and initializes the latter from fused 3D motion evidence;
(3) We propose a practical sparse-view capture and synchronization setup for high-speed real fire acquisition with consumer hardware;
(4) We establish an evaluation protocol for dynamic fire reconstruction combining frame-wise appearance, depth consistency, motion plausibility, and video-level temporal quality, and show that prior GS methods fail under this protocol while our method remains effective. 

\section{Related Work}
\label{sec:related}

Building on a long history of physically based fire simulation~\cite{nguyen2002physically}, we address the inverse problem of recovering 3D geometry and appearance from sparse observations.
Accurate correspondence and tracking form the basis for such reconstruction, linking observed dynamics to coherent 3D representations with Gaussian splats and connecting to broader efforts in long-range point tracking and 4D dynamic reconstruction~\cite{doersch2022tapvid,doersch2023tapir,doersch2024bootstap,harley2022particlevideo,karaev2024cotracker,neoral2024mft,song2024trackeverything,wang2023trackingeverything,koppula2024tapvid3d,ngo2025delta,xiao2024spatialtracker,xiao2025spatialtrackerv2,feng2025st4rtrack,zheng2023pointodyssey,xu2025fourdgt}. The following sections review recent methods in correspondence estimation, view synthesis, and dynamic scene modeling that are related to our approach.

\textbf{Correspondence and Tracking.}
While monocular long-range \emph{3D} tracking remains relatively underexplored, there is a rich literature on tracking in 2D image space via dense or semi-dense correspondence estimation. Classical and modern optical-flow style methods estimate dense motion fields between frames ~\cite{horn1981opticalflow,lucas1981lk,black1993robustflow,brox2004highaccuracyflow,brox2009largedisplacement}, paving the way for multi-frame and occlusion-aware variants~\cite{hur2020selfsupsceneflow,ren2019fusionflow,shi2023videoflow,jiang2021hiddenmotions,jiang2021fewmatchesflow,bargatin2025memfof} including high-resolution multi-frame approaches such as MEMFOF~\cite{bargatin2025memfof}, which we use in our approach. Sparse keypoint and descriptor methods provide more stable, but sparser, correspondences that can be integrated into SfM pipelines~\cite{lowe2004sift,bay2006surf,rublee2011orb,liu2011siftflow,detone2018superpoint,schoenberger2016sfm,schoenberger2016mvs}, enabling long-term tracks in relatively rigid scenes~\cite{bansal2020fourdevents,goli2024romo}. 

Moving from 2D to 3D correspondences, recent work performs geometric matching and reconstruction directly from images: structure-from-motion and MVS pipelines~\cite{schoenberger2016sfm,schoenberger2016mvs}, DUSt3R and MASt3R~\cite{dust3r_cvpr24,mast3r_eccv24}, and the fully integrated MASt3R-SfM system~\cite{duisterhof2025mastrsfm} provide accurate geometric correspondences in static or slowly changing scenes. Depth priors such as Depth Anything~\cite{depth_anything_v1,depth_anything_v2}, Depth Pro~\cite{ICLR2025_bc8b2058}, and Marigold-DC~\cite{viola2024marigolddc} improve per-frame geometry and therefore implicit tracking, but they do not explicitly maintain persistent long-range 3D trajectories. To incorporate geometry, recent work in monocular settings combines optical flow, depth, and rigidity for self-supervised scene-flow estimation and dynamic reconstruction~\cite{hur2020selfsupsceneflow,ranftl2016densemonodepth,li2019frozenpeople,li2021nsff,li2023dynibar,zhang2021consistentdepthmovingobjects,zhang2022structuremotioncasual,luo2020consistentvideodepth,kopf2021robustvideodepth,lu2025align3r}. Feedforward 3D tracking methods estimate trajectories directly in 3D space or focus on rigid reconstruction~\cite{ngo2025delta,xiao2024spatialtracker,xiao2025spatialtrackerv2,feng2025st4rtrack,xu2025fourdgt}. 

\textbf{View Synthesis and Static 3D Reconstruction.}
Neural Radiance Fields (NeRF) introduced continuous volumetric radiance fields for view synthesis from sparse posed images and became the de facto standard for static neural scene representations~\cite{mildenhall2020nerf}. A broad line of work has extended radiance fields to handle video and dynamic content, including free-viewpoint video~\cite{broxton2020immersivelfv,yoon2020dynamicnvs,gao2021dynamicviewsynthesis,li2022neural3dvideo,xian2021stnerf}, articulated humans~\cite{weng2022humannerf,isik2023humanrf,li2022tava,chen2021snarf}, and factorized space--time--appearance representations~\cite{fridovichkeil2023kplanes,wang2022fourierplenoctrees,cao2023hexplane,song2023nerfplayer,du2021neuralradianceflow}.

3D Gaussian Splatting (3DGS) replaces volumetric ray marching with a set of view-dependent anisotropic Gaussians rendered by a differentiable splatting pipeline, achieving real-time, high-quality novel-view synthesis~\cite{Kerbl2023GaussianSplatting}. Some approaches adapt the formulation to 2D Gaussian primitives~\cite{huang2024gs2d}, controllable or editable splats~\cite{Huang_2024_CVPR}, and more general camera and light-transport models with distorted cameras and secondary rays~\cite{wu20253dgut}. These Gaussian-based approaches complement geometric 3D reconstruction methods such as DUSt3R and MASt3R, and their SfM integration~\cite{dust3r_cvpr24,mast3r_eccv24,duisterhof2025mastrsfm}, as well as multi-view dynamic reconstruction systems~\cite{dou2016fusion4d,newcombe2015dynamicfusion,innmann2016volumedeform,zollhoefer2014realtimenonrigid,bozic2020deepdeform}.

\textbf{Dynamic 3D Reconstruction.}
Dynamic scene modeling extends these representations into space-time for 4D reconstruction and animation. Before neural fields, non-rigid reconstruction often relied on RGB-D sensors~\cite{dou2016fusion4d,newcombe2015dynamicfusion,innmann2016volumedeform,zollhoefer2014realtimenonrigid,bozic2020deepdeform} or strong low-rank and NRSfM priors~\cite{bregler2000nonrigid,dai2014nrsfm,novotny2019c3dpo,kumar2017dynamicdense3d,russell2014videopopup}. NeRF-style dynamic reconstructions deform a canonical radiance field using learned deformation fields or higher-dimensional embeddings~\cite{Pumarola2021DNeRF,park2021nerfies,park2021hypernerf,du2021neuralradianceflow,xian2021stnerf,li2021nsff,li2023dynibar,wang2021neuraltrajectoryfields,gao2021dynamicviewsynthesis,yoon2020dynamicnvs,li2022neural3dvideo,song2023nerfplayer}, while articulated and category-specific methods focus on humans and animals~\cite{yang2021lasr,yang2021viser,yang2022banmo,li2019frozenpeople,li2022tava,isik2023humanrf}. Physics-informed neural fields and tomography-based methods reconstruct phenomena such as smoke and flames from sparse observations~\cite{Ihrke2004ImageBased,chu2022physics,gao2025fluidnexus}, but assume controlled multi-view setups or simplified motion models. 
A closely related prior work applies a temporal Laplacian pyramid to blend per-view observations into a free-viewpoint representation of dynamic scenes such as flames~\cite{thonat2021video}. While rendering qualitatively compelling results, this approach does not target a full spatiotemporal reconstruction of the underlying geometry and appearance.

More recently, dynamic 3D perception models aim to couple tracking and reconstruction over long temporal horizons~\cite{wang2025continuous3dperception,feng2025st4rtrack,liu2025modgs,lu2025align3r,goli2024romo,koppula2024tapvid3d,ngo2025delta,xiao2024spatialtracker,xiao2025spatialtrackerv2,xu2025fourdgt}. Building on Gaussian Splatting, several works propose explicitly dynamic Gaussian representations: GS4D and 4D Gaussian Splatting learn 4D Gaussian primitives for real-time photorealistic dynamic view synthesis~\cite{yang2023gs4d,Wu2024FourDGS}; 4D-Rotor Gaussian Splatting introduces rotational dynamics for efficient dynamic scenes~\cite{duan2024rotor4dgs}; spacetime Gaussian feature splatting embeds Gaussians in a joint space--time feature field~\cite{Li_STG_2024_CVPR}; and SC-GS provides sparse controls for editable dynamic scenes~\cite{Huang_2024_CVPR}. Deformable 3D Gaussians and per-Gaussian embedding methods model continuous canonical-space deformations and per-primitive latents~\cite{Yang2024Deformable3DGaussians,Bae2024PerGaussianEmbedding}. Dynamic 3D Gaussians attach per-frame rigid transforms to canonical Gaussians for tracking and view synthesis~\cite{Luiten2024Dynamic3DGaussians}.
Latent-dynamical and monocular setups are further explored by ODE-GS~\cite{Wang2025ODEGS}, FreeTimeGS~\cite{Wang2025FreeTimeGS}, MoSca~\cite{Lei2025MoSca}, DynMF~\cite{kratimenos2024dynmf}, and 4DGT~\cite{xu2025fourdgt}, which respectively introduce latent ODEs, free-time Gaussian primitives, 4D motion scaffolds, motion factorization, and transformer-based 4D Gaussians for real-world monocular videos. RGB-D based methods~\cite{ha2025geometry} produce compelling results on a variety of dynamic real-world scenes but rely on a well-defined surface, which semi-transparent phenomena such as fire and smoke cannot provide.

\begin{figure}
    \centering
    \includegraphics[width=0.7\linewidth]{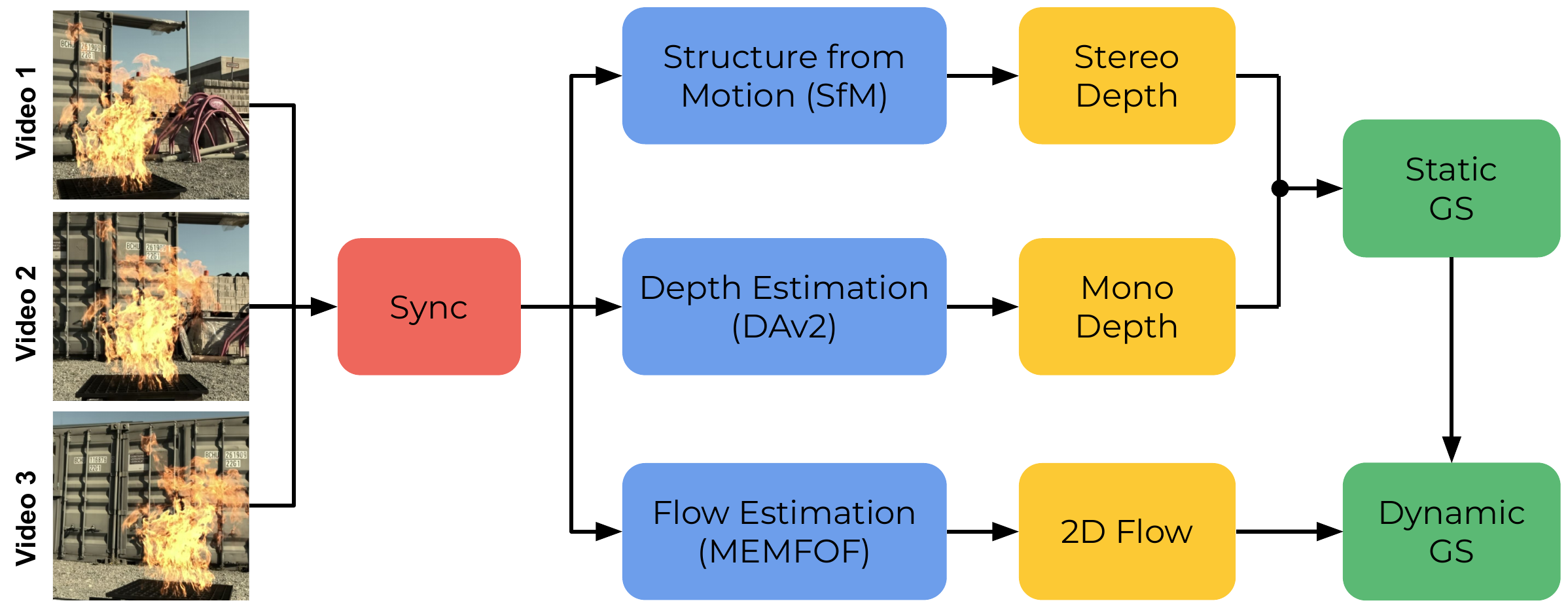}
    \vspace{-1mm}
    \caption{Pipeline overview: Given the video of a fire from three views, we first calibrate and register cameras using COLMAP and estimate dense stereo depth maps (PatchMatchStereo), complemented by aligned monocular depth predictions for regularization. The static background scene is optimized with vanilla 3DGS using this fused depth initialization. Dynamic regions such as fire are reconstructed by projecting dense optical flow into a 3D voxel grid to estimate a volumetric motion field, which initializes our transient flame Gaussians with position, motion, and temporal support. 
    }
    \label{fig:pipeline}
    \vspace{-4mm}
\end{figure}

\section{Methodology}
\label{sec:methods}
As real fire is usually difficult to capture from many viewpoints, our main objective is to use as few cameras as possible for the reconstruction. However, only using a few cameras leads to insufficient view constraints for reconstructing scenes with rendering losses alone. To address this, we separately reconstruct the static scene (background) and the dynamic scene (fire). This enables us to use a view-consistent, rigid representation for the background and a volumetric, dynamic representation for the flames (Fig.~\ref{fig:pipeline}). For the static scene, we use vanilla 3DGS, circumventing the insufficient view constraints by leveraging a dense stereo initialization in combination with monocular depth regularization during optimization. For the dynamic scene, we extend the 3DGS framework with a flow-based initialization that estimates volumetric motion from dense optical flow, enabling the reconstruction of temporally coherent, plausible fire dynamics. All modalities used by our proposed method are depicted in Fig.~\ref{fig:preprocessing}, while Sec.~\ref{supp:decomposition} of the Supp. Mat. visualizes the decomposed static/dynamic scene.

\subsection{Static Scene}

\begin{figure}[t]
    \centering
    \includegraphics[width=\linewidth]{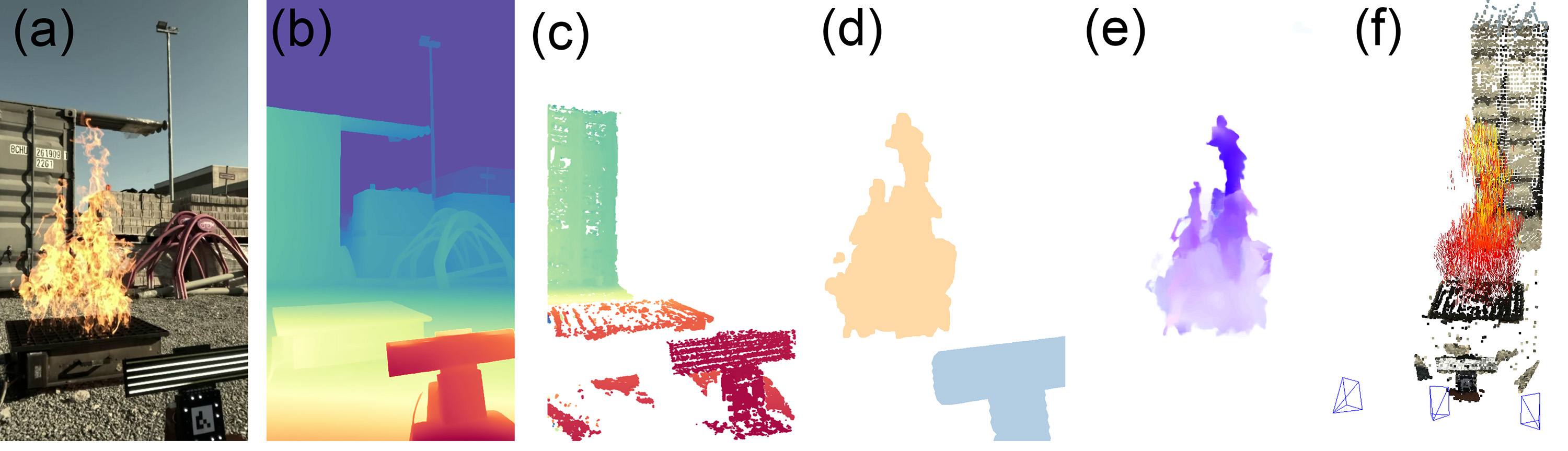}
    \vspace{-4.5mm}
    \caption{Intermediate data used for scene reconstruction: (a) Input RGB frame, (b) monocular depth prediction, (c) depth from structure-from-motion (SfM), (d) segmentation mask of the dynamic region, (e) estimated 2D optical flow, and (f) fused voxelized 3D flow field used to initialize dynamic Gaussians.}
    \label{fig:preprocessing}
    \vspace{-2mm}
\end{figure}

To reconstruct the static scene, we first obtain frames for each view that do not contain any motion. When such frames are not available, we fall back on deriving a background estimate directly from the video sequence. As the flames are actively emitting light, they are usually brighter than their background, allowing us to remove them with a minimum intensity projection along the time dimension (see Sec.~\ref{supp:fire_removal} of the Supp. Mat.). Based on these static frames, we estimate accurate camera poses 
using COLMAP~\cite{schoenberger2016vote}.
To reduce overfitting caused by limited view coverage, we employ two key strategies: (1) a strong point-cloud initialization, and (2) a continuous monocular regularization term.
In order to obtain a dense initial geometry, we apply COLMAP's implementation of PatchMatchStereo~\cite{schoenberger2016vote,bleyer2011patchmatch} to the calibrated and registered static frames and obtain a stereo depth map \(D_{\text{stereo}}\) for each input image.
To initialize regions of the scene observed by at most a single camera, 
we predict monocular depth maps \(D_{\text{mono}}\) using DepthAnythingV2~\cite{depth_anything_v2}. For each camera, we then perform a linear alignment \(D_{\text{mono}}' = a D_{\text{mono}} + b\,\),
where the parameters \(a,b \in \mathbb{R}\) are determined by minimizing the least-squares difference
\begin{equation} \label{eq2}
(a^*, b^*) = \argmin_{a,b\in \mathbb{R}} \| a D_{\text{mono}} + b - D_{\text{stereo}} \|_2^2
\end{equation}
over overlapping regions. We create an initial point cloud by back-projecting the three stereo depth maps. In regions with no stereo information available, we utilize the aligned monocular depth maps to fill the scene (Fig.~\ref{fig:preprocessing}, b). This point cloud then serves as the initialization for a 3DGS reconstruction of the static scene. During training, the aligned monocular depths \(D_{\text{mono}}'\) are incorporated as a regularization term
    \begin{equation} \label{eq3}
    \mathcal{L}_{\text{depth}} = \frac{1}{wh}\sum_{i,j} \left\| D'_{\text{mono}}(i,j) - D_{\text{render}}(i,j) \right\|_1\,,
    \end{equation}
    to prevent degenerate depth solutions and preserve global consistency. 
    The overall loss is given as the weighted sum of the absolute rendering loss \( \mathcal{L}_1 \), the SSIM loss \( \mathcal{L}_\text{SSIM} \), as well as the depth regularization term
    \(
        \mathcal{L} = \lambda_1 \mathcal{L}_1 + \lambda_\text{SSIM} \mathcal{L}_\text{SSIM} + \lambda_\text{depth} \mathcal{L}_\text{depth}\,.
    \)
    We utilize standard parameter values \(\lambda_1 = 0.8\) and \(\lambda_\text{SSIM} = 0.2\). To strengthen the depth regularization ($\lambda_\text{depth}$), we increase the default regularization parameter value by a factor of 100, yielding an exponentially decreasing contribution with an initial value of 100 and a final value of 1~\cite{Kerbl2023GaussianSplatting}. 

\begin{figure}[t]
    \centering
    \includegraphics[width=\linewidth]{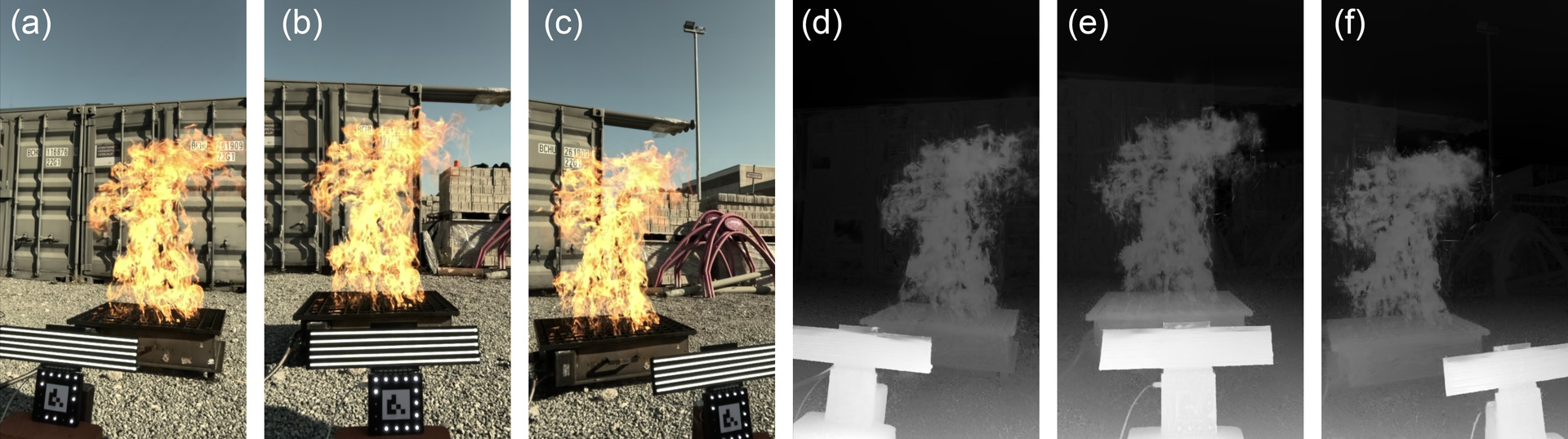}
    \vspace{-4mm}
    \caption{
    Multi-view capture of a fire: we capture the RGB images from the three camera viewpoints (a)--(c) and show the corresponding rendered depth maps (d)--(f). The depth reconstructions demonstrate consistent geometry across views despite strong appearance variations caused by the dynamic flames.}
    \label{fig:view_depth}
    \vspace{-4mm}
\end{figure}

\subsection{Dynamic Scene}

For the dynamic scene, we represent the flame as a set of transient flame Gaussians, i.e., short-lived volumetric primitives whose appearance and motion are optimized only over a finite temporal support. Reconstructing the dynamic component with just dense stereo is not possible for two main reasons: (1) our commodity camera frames are not synchronized, and (2) it does not support obtaining a continuous motion field. 
Instead, to address this, we first estimate an approximate flow field, which is subsequently refined through joint optimization guided by the multi-view camera observations.

We estimate the initial flow field by computing the dense optical flow \( \mathbf{f}_i(u,v) = (f_{i,u}(u,v), f_{i,v}(u,v)) \) for each pixel \( (u,v) \) in the image plane of camera \( i \), where \( \mathbf{f}_i(u,v) \) represents the 2D motion vector at each pixel. The flow vector, which estimates the displacement of each pixel between consecutive frames, is estimated using MEMFOF~\cite{bargatin2025memfof}.
Once the flow is computed, we project all 2D flows onto a voxel grid. 
For each point \( \mathbf{x}\in \mathbb{R}^3 \), the corresponding 2D flows \( \mathbf{f}_i(\mathbf{\pi}_i(\mathbf{x})) \) are back-projected to 3D space
\begin{equation} \label{eq6}
    \mathbf{u}_i(\mathbf{x}) = \mathbf{\pi}_i^{-1}\left( \mathbf{f}_i(\mathbf{\pi}_i(\mathbf{x})) + \mathbf{\pi}_i(\mathbf{x}), \mathbf{x}\right) - \mathbf{x}\,,
\end{equation}
where \( \mathbf{\pi}_i^{-1} ( \ldots, \mathbf{x})\) is the back-projection function to the depth of \( \mathbf{x} \). For each camera, the 3D flow \( \mathbf{F}(\mathbf{x})\in \mathbb{R}^3 \) must adhere to the projection constraint
\begin{equation} \label{eq6a}
    \left(\mathbf{F}(\mathbf{x}) - \mathbf{u}_i(\mathbf{x})\right)^\mathsf{T} \mathbf{u}_i(\mathbf{x}) = 0\,.
\end{equation}
To avoid unstable solutions for under-constrained systems, we apply a Tikhonov regularization with the regularization parameter
\begin{equation} \label{eq6b}
    \tau(\mathbf{x}) = \frac{\tau_0}{m} \sum_i^{m}\|\mathbf{u}_i(\mathbf{x}) \|_2^2 \,,
\end{equation}
where $m$ denotes the number of cameras and $\tau_0$ controls the strength of the regularization.
The 3D flow is finally determined by solving the least-squares system\footnote{\(\mathbf{F}(\mathbf{x})\), \(\mathbf{u}_i(\mathbf{x})\), and \(\tau(\mathbf{x})\) are abbreviated as \(\mathbf{F}\), \(\mathbf{u}_i\), and \(\tau\).}
\begin{equation} \label{eq6c}
\argmin_{\mathbf{F}} 
\left\|
\begin{bmatrix}
\mathbf{u}_1^{\top} \\
\vdots \\
\mathbf{u}_m^{\top}
\end{bmatrix}
\mathbf{F} -
\begin{bmatrix}
\mathbf{u}_1^{\top} \mathbf{u}_1 \\
\vdots \\
\mathbf{u}_m^{\top} \mathbf{u}_m
\end{bmatrix}
\right\|_2^2
+ \tau^2 \|\mathbf{F}\|_2^2\,,
\end{equation}
which is solved via singular value decomposition and completes the preprocessing stage for the dynamic scene.

For modeling the dynamic flames, we chose a constant-velocity  representation inspired by FreeTimeGS~\cite{Wang2025FreeTimeGS}, which assigns three additional parameters to each Gaussian: a time \( t_\mu \), a lifespan \( t_\sigma \), and a linear velocity \( \mathbf{v} \in \mathbb{R}^3\). At any given time, the position of a dynamic Gaussian is determined by \( \mathbf{x}(t) = \mathbf{x}_0 + (t - t_\mu) \mathbf{v} \). Additionally, its opacity is modulated by the temporal modifier \( \sigma(t) = \exp(-\frac{1}{2}(\frac{t - t_\mu}{t_\sigma})^2)\). This representation has two key advantages: (1) it can be directly initialized from a 3D flow field, and (2) it exposes its velocity field explicitly, which enables a variety of downstream tasks for interpreting fire (e.g., robot navigation).

Given the dynamic representation, for each position \( \mathbf{x} \) in the voxel grid at time \( t \), we assign a Gaussian with velocity \( \mathbf{F}(\mathbf{x}) \) and time \( t_\mu = t \). The Gaussian's lifetime is initialized as four times the frame time for smooth motion and to prevent overfitting to single frames. To account for voxel discretization, we introduce a random offset in the position of the Gaussian, sampled uniformly from the voxel volume, with \( \Delta \mathbf{x} \sim \mathrm{Uniform}([-s/2, s/2]^3) \), where \( s \) is the voxel size, to avoid moiré patterns. A multi-view capture of fire is shown in Fig.~\ref{fig:view_depth}.

To further mitigate overfitting of the dynamic Gaussians to the static background, during training we compare the opacity of the dynamic Gaussians \( A_{\text{dyn}} \) to a precomputed motion mask \(M\) and compute an additional \textit{alpha loss} as \(\frac{1}{wh}\sum A_{\text{dyn}} (1 - M)\) with a weight of \(0.1\), which discourages placing dynamic Gaussians in regions where no flames have been detected in the input video.

\subsection{Visual Synchronization}
To capture high-frequency phenomena such as fire from multiple viewpoints, precise temporal synchronization is essential. However, most consumer-grade hardware lacks native support for high-frame-rate synchronization. To address this, we design a low-cost synchronization system based on a microcontroller. Specifically, we use an ESP32 microcontroller to control the logic-level outputs of \( 21 \) pins. These pins are divided into two groups: five pins generate a sub-frame synchronization pattern, while the remaining \( 16 \) pins manage an incrementing frame counter. The frame counter is displayed using \( 15 \) white LEDs, which represent the frame number in Gray code. An additional LED is employed to resolve the inherent ambiguity of the Gray code, preventing frame errors that could otherwise reach up to one frame. The synchronization pattern is shown in \cref{fig:preprocessing}a.

To achieve sub-frame timing with microsecond-level precision, we exploit the pronounced rolling shutter effect present in most CMOS consumer cameras by sequentially toggling a series of five COB LED strips. The rolling shutter becomes clearly visible by observing the partially illuminated LED strips. For a fully characterized sensor, detecting the image points where the LED strip brightness changes allows us to directly determine the exposure time. The temporal precision is directly dependent on the row time \( t_{\text{row}} \) and the spatial accuracy \( \Delta b \) in detecting the change in LED strip brightness. For typical values of \( t_{\text{row}} \approx \qty{3}{\micro\second} \) and \( \Delta b \approx \qty{5}{\pixel} \), we achieve a temporal precision of approximately \( \qty{15}{\micro\second} \). For an explicit representation of the rolling shutter within our proposed method, we refer to Sec.~\ref{supp:rolling-shutter} of the Supp. Mat.

\begin{figure*}[t]
    \centering
    \includegraphics[width=1.0\textwidth]{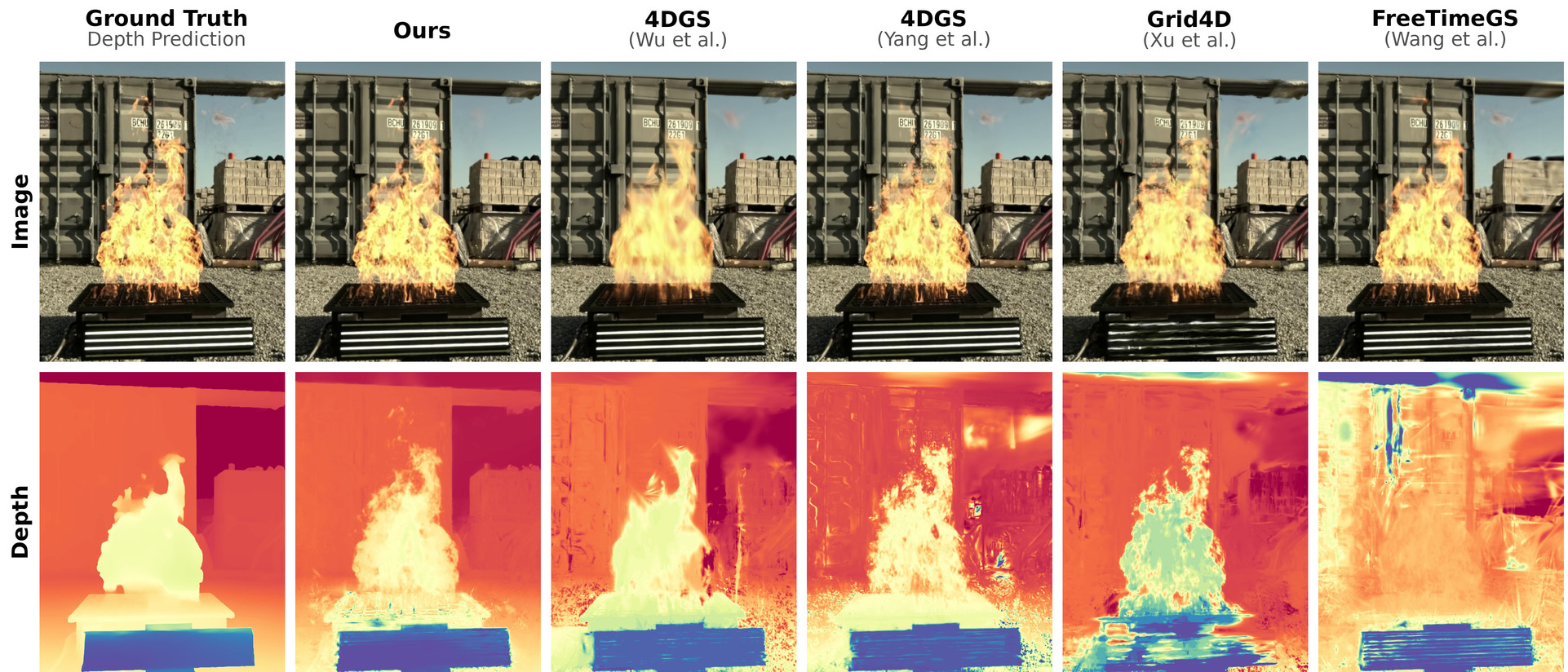}
    \vspace{-3mm}
    \caption{Qualitative comparison of our method against 4DGS~\cite{Wu2024FourDGS}, 4DGS~\cite{yang2023gs4d}, Grid4D~\cite{10.5555/3737916.3741850}, and FreeTimeGS~\cite{Wang2025FreeTimeGS}. We provide the ground truth image as well as the rendered results from all methods. To evaluate the plausibility of the rendered depth, we provide a monocular depth estimation for comparison~\cite{depth_anything_v2}. While other methods struggle with degradation of visuals or depth, our method provides rendered images of high visual quality while avoiding depth inaccuracies.}
    \vspace{-4mm}
    \label{fig:compositions}
\end{figure*}

\subsection{Velocity Metrics}
In reconstruction tasks, the most commonly used metrics such as PSNR, LPIPS, and SSIM are used to quantitatively assess similarity of 2D images. However, these metrics cannot directly be applied to assess similarity of dynamic 3D scenes. For dynamic similarity evaluation, some methods render 2D optical flow from the scene and evaluate this against estimated optical flow~\cite{zhu2024motiongs}, while others observe the motion of the Gaussian centers and compare these against known trajectories~\cite{Luiten2024Dynamic3DGaussians}. Neither approach is feasible for our case. Comparing against 2D optical flow would introduce a circular dependency into our pipeline, and would rely on estimations of the real flow. For the second approach, we determined that tracking the center position of the Gaussians is insufficient to capture their full dynamics, as each Gaussian represents a spatial distribution and varying opacity and shape. 
Therefore, we propose a novel metric for assessing similarity of 3D GS reconstructed scenes of transient phenomena -- in our case flames. Specifically, we employ synthetic scenes to evaluate the motion reconstruction capability of our model, as these provide, in addition to the dynamics, the underlying mechanism -- the velocity field -- to be used for evaluation, which we refer to as the density-weighted volumetric evaluation protocol (DVE). For every scene, we obtain the \(N\) optimized Gaussians with their center position \(\mathbf{x}\), opacity \(\alpha\), scale \( \mathbf{s} \), velocity \(\mathbf{v}\), and rotation \(R\). Based on this, we compute the Gaussian's weight \(w = s_x s_y s_z \alpha\). Given the normalized Gaussian probability distribution \(P(\mathbf{x}, \mathbf{s}, R)\) and the underlying ground truth velocity field \(V\), we approximate the expected value of the ground truth velocity function \(\mathbf{\bar{v}} = E_P(V)\) by employing a third-order Gauss-Hermite quadrature. We capture the overall similarity of the two motion fields using a relative L2 metric
\begin{equation}
    \text{L2} = \sqrt{\frac{\sum_{n=1}^N w_n \lVert\mathbf{v}_n - \mathbf{\bar{v}}_n\rVert^2}{\sum_{n=1}^N w_n \lVert\mathbf{\bar{v}}_n\rVert^2}} \ , 
\end{equation}
and measure directional alignment via weighted volumetric cosine similarity
\begin{equation}
    \text{CosSim} = \frac{\sum_{n=1}^N w_n \frac{\mathbf{v}_n \cdot \mathbf{\bar{v}}_n}{\lVert\mathbf{v}_n\rVert \lVert \mathbf{\bar{v}}_n\rVert}}{\sum_{n=1}^N w_n} \ .
\end{equation}
Together, these metrics yield a density-aware evaluation protocol capturing the quantitative accuracy and directional fidelity of the reconstructed motion field.

\section{Dataset}
We record a real-world dataset of fire sequences using three \emph{GoPro Hero 13 Black} cameras arranged in a fixed multi-view setup. Each captures at \qty{400}{\hertz}, which provides a high-frequency temporal resolution necessary for modeling the rapid motion and radiative changes of fire. We use three cameras as this enables multi-view cross-validation for robust COLMAP points, which provide a fair initialization for the baseline methods. Each sequence spans approximately \qty{15}{\second}. Furthermore, we captured fire against different backgrounds to evaluate reconstruction robustness under varying lighting and texture conditions.

To capture diverse flame dynamics and appearance, we include several fire materials such as propane jets, burning wood, gasoline, cardboard, and paper. In total, we captured 17 videos of fire for each camera. For each of these scenes, we created a subsequence containing 100 frames per camera. This setup provides a controlled yet challenging dataset for testing our monocular and multi-view fire reconstruction approach. Code and data will be released with the paper. 

To compute the DVE similarity between reconstruction and ground truth and compare our method against existing baselines in the GS domain, we simulated 3 fire scenes in \textit{Mantaflow}. Specifically, we created a setup of three cameras observing fires, placed within three \textit{Blender} scenes. Aside from 50 ray-traced frames per camera over \qty{2}{\second} of simulation, this simulation provides us with a ground truth velocity field with a resolution of $640^3$, which enables our DVE similarity assessment. In total, we generated 150 synthetic velocity fields. For further details on both datasets, refer to Sec.~\ref{supp:dataset-details} of the Supp. Mat.
\section{Results} \label{sec:results}
To evaluate our proposed method on the real-world fire dataset, we held out every 8th frame for validation. As shown in \cref{fig:compositions}, our method reconstructs the captured scenes with high visual fidelity while preserving the spatiotemporal evolution of the flames. Qualitative results for the additional scenes across a wide range of materials, including wooden logs, cardboard, and paper, are included in Sec.~\ref{supp:results-details} of the Supp. Mat., while additional videos clarify the recovered spatiotemporal behavior, in particular the temporal coherence of the flames and the failure modes of the baseline methods (see Sec.~\ref{supp:novel-views} of the Supp. Mat.).

\begin{figure*}[t]
    \centering
    \includegraphics[width=1.0\textwidth]{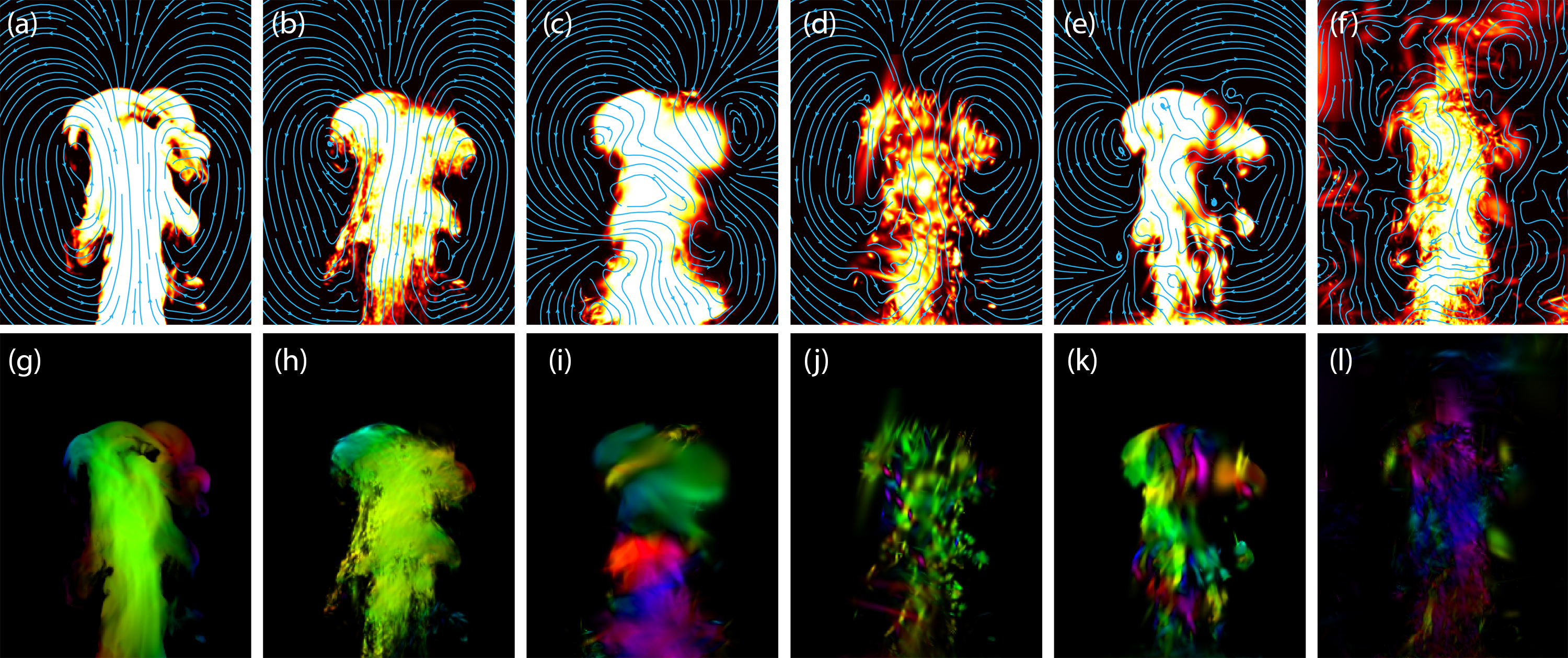}
    \vspace{-4mm}
    \caption{Stream plot comparisons between maximum projected Gaussian velocities of simulated ground truth~(a), our reconstruction~(b), Grid4D~\cite{10.5555/3737916.3741850}~(c), 4DGS~\cite{yang2023gs4d}~(d), 4DGS~\cite{Wu2024FourDGS}~(e), FreeTimeGS~\cite{Wang2025FreeTimeGS}~(f). The streamlines indicate the flow of the Gaussians and the colors indicate their density. The second row (g--l) depicts the color-coded Gaussian velocities.}
    \vspace{-6mm}
    \label{fig:simulation-flow}
\end{figure*}

For a quantitative evaluation, we apply 4DGS~(Wu et al.)~\cite{Wu2024FourDGS}, 4DGS~(Yang et al.)~\cite{yang2023gs4d}, Grid4D~\cite{10.5555/3737916.3741850}, and FreeTimeGS~\cite{Wang2025FreeTimeGS} as established baseline methods. Aside from FreeTimeGS, which applies its own RoMa initialization~\cite{edstedt2024roma}, all other baseline methods were initialized from COLMAP points. We used the author-provided hyperparameter configurations for the DyNeRF \emph{flame salmon} scene. To compare the overall visual quality, we report PSNR, SSIM, and LPIPS. For the more local metrics PSNR and SSIM, we mask out the synchronization pattern. We limit the quantitative comparison to GS-based baselines because they are the only class of methods that operate in the same efficiency regime targeted in this work. Furthermore, we provide additional masked variants PSNR\textsubscript{flame} and SSIM\textsubscript{flame} that only evaluate reconstruction quality in those regions of the image that contain motion. To evaluate the plausibility of the rendered depth of the flame, we compare it against a predicted monocular depth after linear alignment and report the score as RMSE\textsubscript{depth}. A central methodological point of our evaluation is that frame-wise image metrics alone are insufficient for dynamic fire reconstruction: methods with partially competitive PSNR can still fail catastrophically in 3D structure and temporal behavior. Taken together, the image, depth, video, and velocity metrics allow us to distinguish between methods that merely match frame appearance and methods that recover a temporally and geometrically meaningful 4D reconstruction. For a novel-view evaluation of our method on synthetic scenes with spatially held-out views, we refer to Sec.~\ref{supp:spatial-holdout} of the Supp. Mat.
As shown in Tab.~\ref{tab:results-real}, our proposed method\footnote{This evaluation of our method includes a correction in the initialization of the dynamic Gaussians, which in the ECCV-published version contained a temporal offset. This affected only real-world scenes reconstructed by our method and its ablations. The metrics in \cref{tab:results-real} and Supp.~Tab.~\ref{tab:per_scene_results}--\ref{tab:cgvqm_results} have been updated accordingly.} outperforms all baselines in all visual metrics for both the flame and the full image, as well as in the plausibility of the rendered depth. FreeTimeGS achieves second-best flame reconstruction results, but only at the cost of a high degree of depth degradation. 4DGS~\cite{yang2023gs4d} achieves second-best full-frame visual metrics. Further video metrics yield similar results and are provided in Sec.~\ref{supp:video-metrics} of the Supp. Mat.

\begin{table*}[t]
\centering
\renewcommand{\arraystretch}{1.2}
\caption{Quantitative comparison and ablations for the real-world dataset. Our method outperforms all baselines in the visual metrics for the reconstruction of the fire as well as in the accuracy of the predicted flame depth.}
\vspace{-2mm}
\scalebox{0.8}{
\begin{tabular}{lcccccc}
\hline
\textbf{Method} & 
\textbf{PSNR\textsubscript{flame}}~($\uparrow$) & 
\textbf{SSIM\textsubscript{flame}}~($\uparrow$) & 
\textbf{RMSE\textsubscript{depth}}~($\downarrow$) & 
\textbf{LPIPS}~($\downarrow$) & 
\textbf{PSNR}~($\uparrow$) & 
\textbf{SSIM}~($\uparrow$)\\
\hline
\textbf{Ours} & 
\( \first{30.53 \pm 2.33} \) & 
\( \first{0.897 \pm 0.041} \) & 
\( \first{0.090 \pm 0.039} \) & 
\( \first{0.030 \pm 0.008} \) & 
\( \first{37.54 \pm 2.38} \) & 
\( \first{0.973 \pm 0.009} \) \\
\textbf{4DGS~\cite{yang2023gs4d}} & 
\(\third{23.28 \pm 1.76}\) & 
\(\third{0.728 \pm 0.069}\) & 
\(\third{0.113 \pm 0.045}\) & 
\(\second{0.053 \pm 0.020}\) & 
\(\second{32.41 \pm 2.86}\) & 
\(\second{0.945 \pm 0.021}\)\\
\textbf{4DGS~\cite{Wu2024FourDGS}} & 
\(20.99 \pm 1.14\) & 
\(0.601 \pm 0.054\) & 
\(\second{0.102 \pm 0.036}\) & 
\(\third{0.071 \pm 0.025}\) & 
\(\third{31.88 \pm 3.12}\) & 
\(\third{0.933 \pm 0.025}\)\\
\textbf{FreeTimeGS} & 
\(\second{24.48 \pm 3.48}\) &
\(\second{0.756 \pm 0.132}\) &
\(0.131 \pm 0.042\) &
\(0.241 \pm 0.131\) &
\(25.62 \pm 1.83\) &
\(0.794 \pm 0.109\) \\
\textbf{Grid4D} &
\(18.52 \pm 1.47\) &
\(0.497 \pm 0.068\) &
\(0.225 \pm 0.060\) &
\(0.086 \pm 0.025\) &
\(29.16 \pm 3.33\) &
\(0.909 \pm 0.032\) \\
\hline \hline
\textbf{Full Method} & 
\( \second{30.53 \pm 2.33} \) & 
\( \second{0.897 \pm 0.041} \) & 
\( \second{0.090 \pm 0.039} \) & 
\( \first{0.030 \pm 0.008} \) & 
\( \second{37.54 \pm 2.38} \) & 
\( \first{0.973 \pm 0.009} \) \\
\textbf{NoSync} & 
\( \third{30.20 \pm 2.17} \) & 
\( \third{0.893 \pm 0.041} \) & 
\( \third{0.091 \pm 0.039} \) & 
\( \third{0.031 \pm 0.008} \) & 
\( \third{37.43 \pm 2.43} \) & 
\( \third{0.972 \pm 0.009} \)\\
\textbf{RandFlow} & 
\( 27.55 \pm 2.85 \) & 
\( 0.823 \pm 0.082 \) & 
\( \first{0.088 \pm 0.040} \) & 
\( 0.035 \pm 0.013 \) & 
\( 36.26 \pm 3.35 \) & 
\( 0.965 \pm 0.018 \)\\
\textbf{JointOpt} & 
\( 29.48 \pm 2.28 \) & 
\( 0.889 \pm 0.042 \) & 
\( 0.104 \pm 0.038 \) & 
\( 0.055 \pm 0.013 \) & 
\( 34.48 \pm 1.61 \) & 
\( 0.954 \pm 0.009 \)\\
\textbf{NoMask} & 
\( \first{30.78 \pm 2.42} \) & 
\( \first{0.900 \pm 0.042} \) & 
\( 0.097 \pm 0.039 \) & 
\( \first{0.030 \pm 0.008} \) & 
\( \first{37.87 \pm 2.44} \) & 
\( \first{0.973 \pm 0.009} \)\\
\hline
\end{tabular}
}
\label{tab:results-real}
\vspace{-2mm}
\end{table*}

\begin{table*}[t]
\centering
\caption{Quantitative comparison for the synthetic fire scene. Our method yields the most physically plausible reconstruction with well aligned velocities, while the baselines only show a weak directional correlation with the ground truth.}
\vspace{-2mm}
\renewcommand{\arraystretch}{1.2}
\setlength{\tabcolsep}{4.5pt} %
\scalebox{0.79}{
\begin{tabular}{lcc @{\hspace{1.0cm}} lcc} %
\hline
\textbf{Method} & \textbf{L\textsubscript{2}}~($\downarrow$) & \textbf{CosSim}~($\uparrow$) & \textbf{Ablation} & \textbf{L\textsubscript{2}}~($\downarrow$) & \textbf{CosSim}~($\uparrow$) \\
\hline
\textbf{Ours}               & \first{\( 0.825 \pm 0.058 \)}  & \first{\( 0.744 \pm 0.032 \)}  & \textbf{Full Method} & \second{\( 0.825 \pm 0.058 \)} & \second{\( 0.744 \pm 0.032 \)} \\
\textbf{4DGS (Yang et al.)} & \second{\( 1.045 \pm 0.194 \)} & \second{\( 0.225 \pm 0.123 \)} & \textbf{NoSync}      & \third{\( 0.838 \pm 0.058 \)}  & \third{\( 0.736 \pm 0.034 \)} \\
\textbf{4DGS (Wu et al.)}   & \( 2.218 \pm 0.276 \)          & \( 0.036 \pm 0.010 \)          & \textbf{RandFlow}    & \( 1.519 \pm 0.026 \)          & \( 0.338 \pm 0.050 \) \\
\textbf{FreeTimeGS}         & \third{\( 1.164 \pm 0.042 \)}  & \third{\( 0.194 \pm 0.114 \)}  & \textbf{JointOpt}    & \first{\( 0.820 \pm 0.063 \)}  & \first{\( 0.749 \pm 0.033 \)} \\
\textbf{Grid4D}             & \( 2.262 \pm 1.470 \)          & \( 0.154 \pm 0.014 \)          & \textbf{NoMask}      & \( 0.850 \pm 0.059 \)          & \( 0.697 \pm 0.035 \) \\
\hline
\end{tabular}
}
\label{tab:simulation}
\vspace{-5.5mm}
\end{table*}

To evaluate the reconstructed motion, we apply our proposed DVE protocol to the synthetic scenes. We report the weighted volumetric relative L2 metric and cosine similarity. The results are reported in~\cref{tab:simulation}. While the comparison methods mostly fail in reconstructing physically plausible motion, our method shows a comparatively high similarity between the reconstruction and the ground truth flow field. Especially for the directional similarity, the baselines show largely weak alignment with the true flow field, while our velocities are mostly correctly aligned. This is further visualized in \cref{fig:simulation-flow}, which depicts an X-slice of the reconstructed fire simulation.

To demonstrate the impact of the individual components of our approach, we include several ablations over the timing, the flow field initialization, the alpha loss, and the separation of static and dynamic optimization. 
The evaluation is reported in \cref{tab:results-real} for the real-world scenes and in \cref{tab:simulation} for the synthetic scenes. 
Introducing timing offsets of half a frame-time (\textit{NoSync}) results in reduced visual quality in the flames for the real-world experiments and a reduction in the accuracy of the reconstructed motion field for the simulated scenes. 
An even stronger effect is observable when replacing our flow-based initialization with a random initialization of the dynamic Gaussians (\textit{RandFlow}), which results in substantial visual and physical degradation. This indicates that without proper initialization, the in-training optimization of Gaussian velocities is insufficient to find a faithful approximation of the underlying volumetric flow just from photometric supervision.
Similarly, omitting the alpha loss causes significant degradation of the reconstructed motion fields while improving the overall visual results (\textit{NoMask}). 
The last ablation thus demonstrates a delicate trade-off between visual fidelity and physical accuracy: applying the alpha loss removes the method's ability to place dynamic Gaussians in static parts of the scene, thus improving the physical accuracy while simultaneously reducing the visual capacity.
In contrast, \textit{JointOpt}'s degradation is mostly concentrated in full-frame fidelity, indicating that this ablation primarily shows the static-pretraining's contribution to the full-frame visuals rather than being a physical-visual trade-off.

\section{Discussion and Limitations}
Our results show that a dynamic Gaussian representation can be adapted to a regime where existing methods fail: sparse-view reconstruction of transient, emissive, semi-transparent flames. By combining static stereo-monocular depth fusion with dynamic flow-based initialization and masked optimization, our method recovers reconstructions that are not only visually faithful but also substantially more consistent in depth and motion. The resulting pipeline enables efficient reconstruction of high-speed fire events using consumer-grade hardware and lightweight synchronization.
Despite these strengths, several limitations remain. For one, the reliance on dense optical flow introduces inaccuracies in regions of strong emissive variation or low texture, which can propagate into the 3D motion field. The failure mode of flames missed by the flow estimator during the flow-based initialization cannot be recovered during the dynamic optimization. Second, the current linear velocity model may not fully capture the complex, turbulent motion of real flames, leading to temporal blurring or oversmoothing. The specific failure modes are discussed further in Sec.~\ref{supp:failure-cases} of the Supp. Mat.

Another theoretical limitation of our method is the linear transmittance rasterizer. In our method, we rely on the vanilla 3DGS rasterizer that approximates transmittance linearly over a more physically accurate exponential attenuation. Initial experiments showed that due to our specific initialization our dynamic scene consists of a large quantity of small Gaussians (\(\approx 300\) per ray), so that the linear approximation is especially close to the exponential model. Thus, the added complexity of an exponential rasterizer is not justified for these fire scenes. For further details we refer to Sec.~\ref{supp:exp-transmittance} of the Supp. Mat.

Additionally, separating the static and dynamic components assumes minimal background-foreground interaction, which may not hold for large-scale or smoke-rich scenes. For reconstructing other volumetric phenomena with our flow-based pipeline, we refer to Sec.~\ref{supp:other-domains} of the Supp. Mat. Finally, while our LED-based synchronization mitigates temporal offsets, sub-millisecond inaccuracies can still introduce artifacts at very high motion speeds.

\section{Conclusion}
We presented a method for reconstructing the dynamics of real fire from high-speed multi-view video using 3D Gaussians. Our approach enables temporally coherent reconstructions of flames with substantially improved motion alignment and depth consistency, even under sparse-view capture and consumer-camera timing offsets.
As future work, we plan to integrate physics-based priors for combustion dynamics to improve the photometric modeling of emissive materials. This will enable more adaptive temporal parameterizations of Gaussians and incorporate learning-based flow estimation directly in 3D space. Furthermore, this would allow us to extend the system toward fully monocular dynamic reconstruction for greater fidelity and robustness. Another promising direction is to explore the joint optimization of per-camera timing offsets, thus eliminating the need for our custom LED synchronization setup.

\section*{Acknowledgments}
We thank Marvin Voigt, Anton Wagner, and Helge Wrede for their assistance in executing the experiments and Michael Lütten for providing the hardware used for the propane fire setup. Furthermore, the authors acknowledge the financial support of Catalyst: Leaders Julius von Haast Fellowship (23-VUW-019-JVH). Sören Pirk wishes to acknowledge the European Research Council (ERC) who partially funded this research through the ERC Consolidator Grant WildfireTwins (Grant agreement ID: 101170158).

\bibliographystyle{splncs04}
\bibliography{main}

\clearpage
\appendix
\renewcommand{\theHsection}{supp.\Alph{section}}
\renewcommand{\topfraction}{0.9}
\renewcommand{\bottomfraction}{0.8}
\renewcommand{\textfraction}{0.07}
\renewcommand{\floatpagefraction}{0.75}
\setcounter{topnumber}{3}
\setcounter{totalnumber}{5}
\markright{Gaussians on Fire: Supplementary Material}%
\pdfbookmark[1]{Supplementary Material}{suppmat}
\begin{center}
  {\Large\bfseries\boldmath\pretolerance=10000 Supplementary Material\par}
  \vskip .3cm
  {\large\bfseries\boldmath\pretolerance=10000 Gaussians on Fire:\\High-Frequency Reconstruction of Flames\par}
\end{center}
\vskip .8cm
\noindent This document provides additional material complementing the main paper. We include qualitative visualizations of reconstructed scene trajectories and our reconstruction pipeline, extended implementation details, a detailed description of the captured fire dataset, our rolling shutter implementation, comprehensive per-scene results, scene decomposition details, a spatial holdout validation, tests on other volumetric phenomena, a discussion on the linear transmittance model, and an overview of failure cases.

\section{Fire Removal}
\label{supp:fire_removal}
\begin{figure}[htbp]
    \centering
    \includegraphics[width=\linewidth]{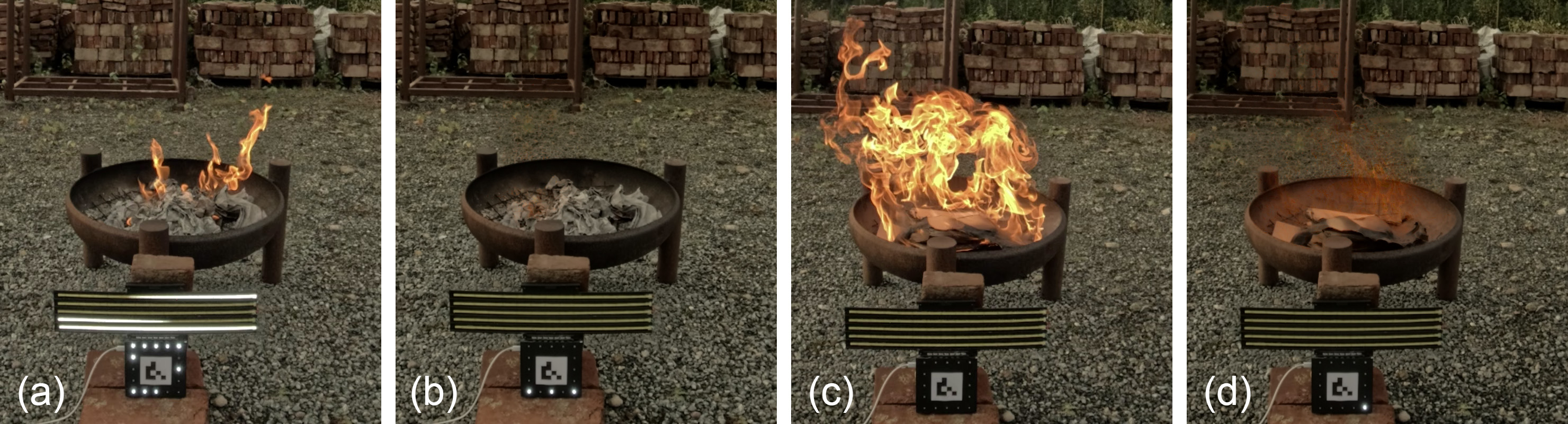}
    \vspace{-6mm}
        \caption{Fire removal for dynamic scenes: we stack all frames along the sequence dimension and compute the minimum intensity projection over this axis to obtain an image of the static scene with most of the fire removed. We show a scene with a small (a,b) and larger flame (c,d) before and after this process.}    
    \label{fig:fire_removal}
    \vspace{-4mm}
\end{figure}

The fire in our experiments is mostly emissive and raises the brightness of each image pixel where it is present. For a short video sequence with a static camera and the only motion being emissive flames, we can thus perform a minimum intensity projection over the time axis to remove the fire from the scene and reveal the background behind the flames. \cref{fig:fire_removal} demonstrates this procedure. As shown in this example, even for large flames this method removes most of the flames very effectively, enabling a robust reconstruction of the static background.

\section{Scene Decomposition}
\label{supp:decomposition}
\begin{figure}[htbp]
  \centering
  \newcommand{\decomppanel}[2]{%
    \begin{minipage}[t]{0.31\linewidth}%
      \centering
      {\scriptsize\bfseries #2}\\[0.5mm]
      \includegraphics[width=\linewidth]{#1}%
    \end{minipage}%
  }
  \decomppanel{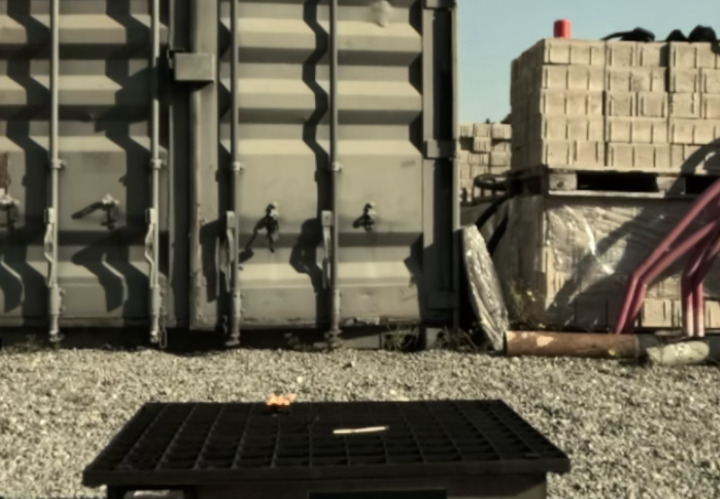}{Static}\hfill
  \decomppanel{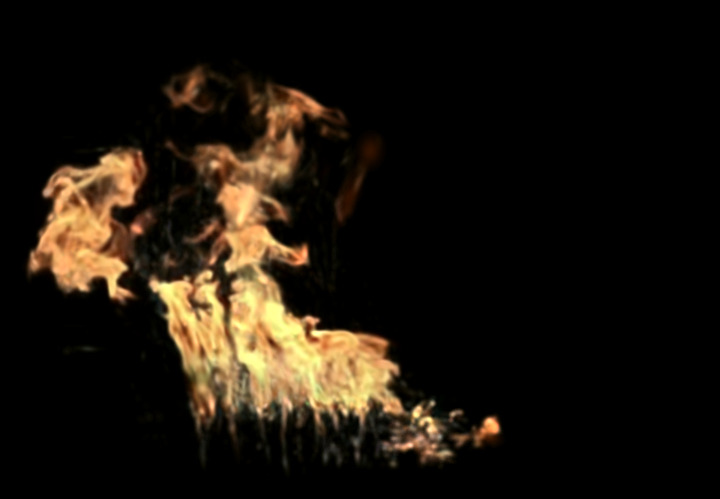}{Dynamic}\hfill
  \decomppanel{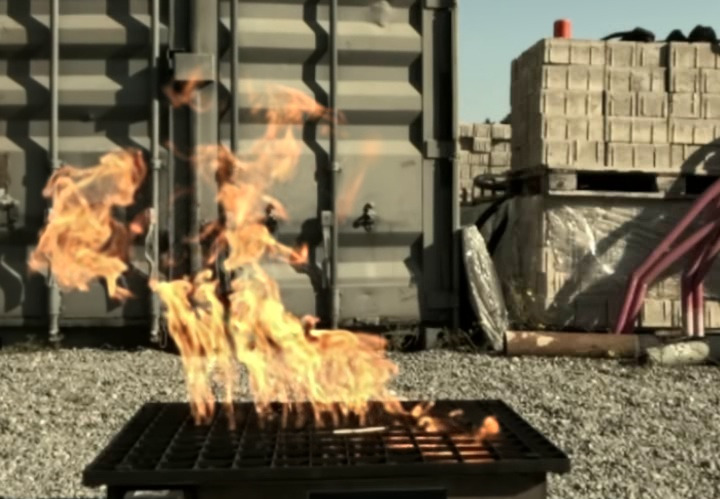}{Combined}
  \caption{Static/dynamic scene decomposition.}
  \label{fig:decomposition}
\end{figure}

In \cref{fig:decomposition}, we visualize the different components of an exemplary scene. \emph{Static} shows the result of the fire-removal process (with a small remaining fire artifact), \emph{Dynamic} shows only the dynamic Gaussians, while \emph{Combined} shows the rendered image of the full scene. The full-sequence videos on the project website\footnote{\url{https://jna-358.github.io/gaussians_on_fire/}} contain per-frame dynamic-only renderings, which provide a more detailed view of the reconstructed flames across the whole sequence.

\section{Rolling Shutter}
\label{supp:rolling-shutter}
Since most high-frame-rate consumer cameras rely on CMOS sensors with significant readout times, the resulting images are prone to rolling shutter distortion. In our experiments, we observed that for our pipeline, not explicitly handling rolling shutter distortions did not reduce the reconstruction quality in any metric. At the high frame rates, the inter-frame motion is comparatively small, which enables the reconstruction method to embed the rolling shutter distortion within the scene geometry, which removes the need for an explicit representation of rolling shutter effects.

However, to also handle scenes with much faster motion, we provide a simple extension to our method, which explicitly models rolling shutter effects. 
Given the sensor specifications and the undistortion parameters, we compute a non-linear delay function \( t(\mathbf{p}) \) for each image point \( \mathbf{p} \). 
For a dynamic Gaussian observed at \( \mathbf{p}_0 \) at the start of the frame (\( t = 0 \)), the projection of its linear worldspace motion itself is not linear, so we approximate its 2D motion as \( \mathbf{p}(t) \approx \mathbf{p}_0 + t \, \mathbf{v}_{\text{image}} \), where \( \mathbf{v}_{\text{image}} \) is the velocity in the image plane.

To estimate the temporal offset for this Gaussian, we locally approximate the nonlinear delay function as \( t(\mathbf{p}) \approx t(\mathbf{p_0}) + \nabla t \cdot (\mathbf{p} - \mathbf{p}_0) \). Combining these two linear approximations, we find that the Gaussian is captured with a delay given by
\begin{equation} \label{eq7}
t = \frac{t(\mathbf{p}_0)}{1 - \nabla t \cdot \mathbf{v}}\,.
\end{equation}

For our setup, we observed a readout time of \qty{2.85}{\micro \second}, so for a maximum horizontal screen space motion of \qty{50}{\pixel} per frame, this gives \( \nabla t \cdot \mathbf{v} \approx 5.5 \cdot 10^{-2} \). For such motions, we can approximate the temporal offset directly as \( t(\mathbf{p}_0) \) in our rendering pipeline, yielding only sub-pixel approximation errors.

\section{Dataset Details}
\label{supp:dataset-details}
To demonstrate the capabilities of our method, we need to show its results with respect to (1) the visual fidelity of the reconstruction, and (2) its physical plausibility. For the former we captured an extensive real-world dataset using three visually synchronized cameras, and for the latter, we ray-traced several simulated fire scenes and obtained their underlying velocity fields. This section gives an overview of the datasets created across both domains.

\subsection{Real-World Dataset}
\label{supp:dataset-real}
\begin{figure*}
    \centering
    \includegraphics[width=\linewidth]{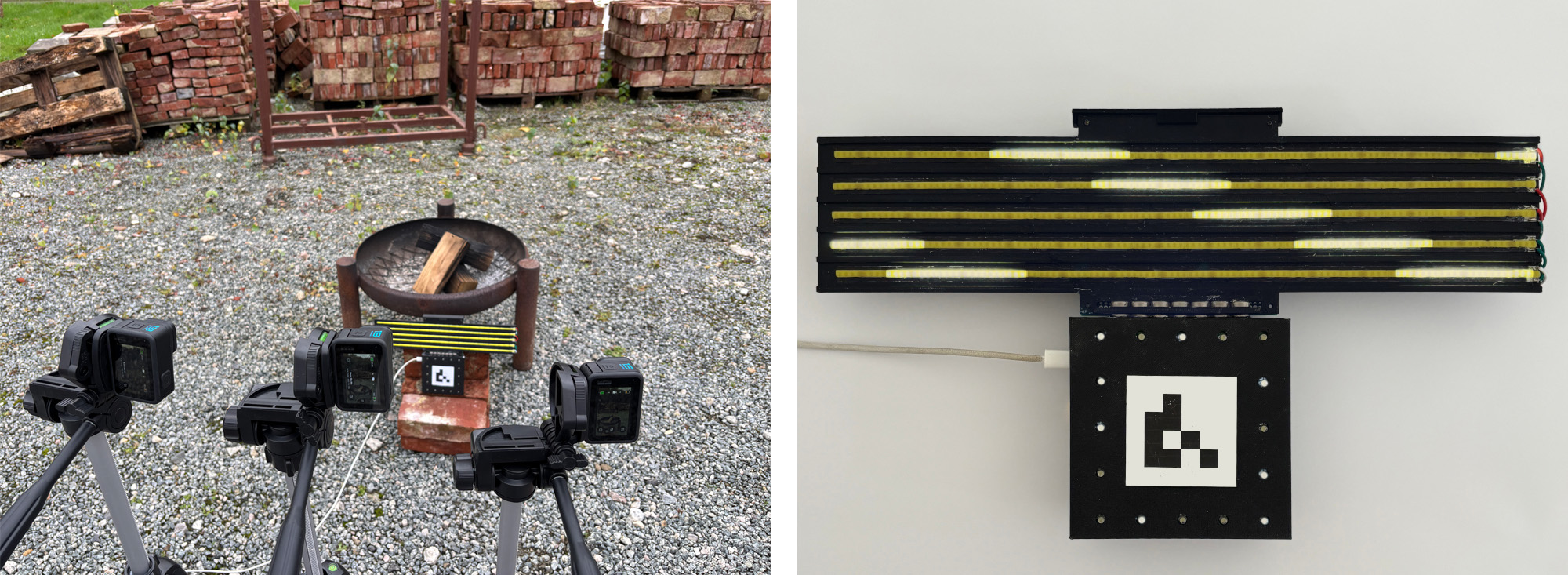}
    \vspace{-6mm}
    \caption{Real-world capturing setup. The left image depicts the camera and fire setup, while the right image shows the custom hardware used for visual synchronization of the cameras. The partially lit LEDs within its top section clearly depict the rolling shutter effect used for sub-frame synchronization.
    }
    \vspace{-5mm}
    \label{fig:setup}
\end{figure*}

\begin{figure*}
    \centering
    \includegraphics[width=\linewidth]{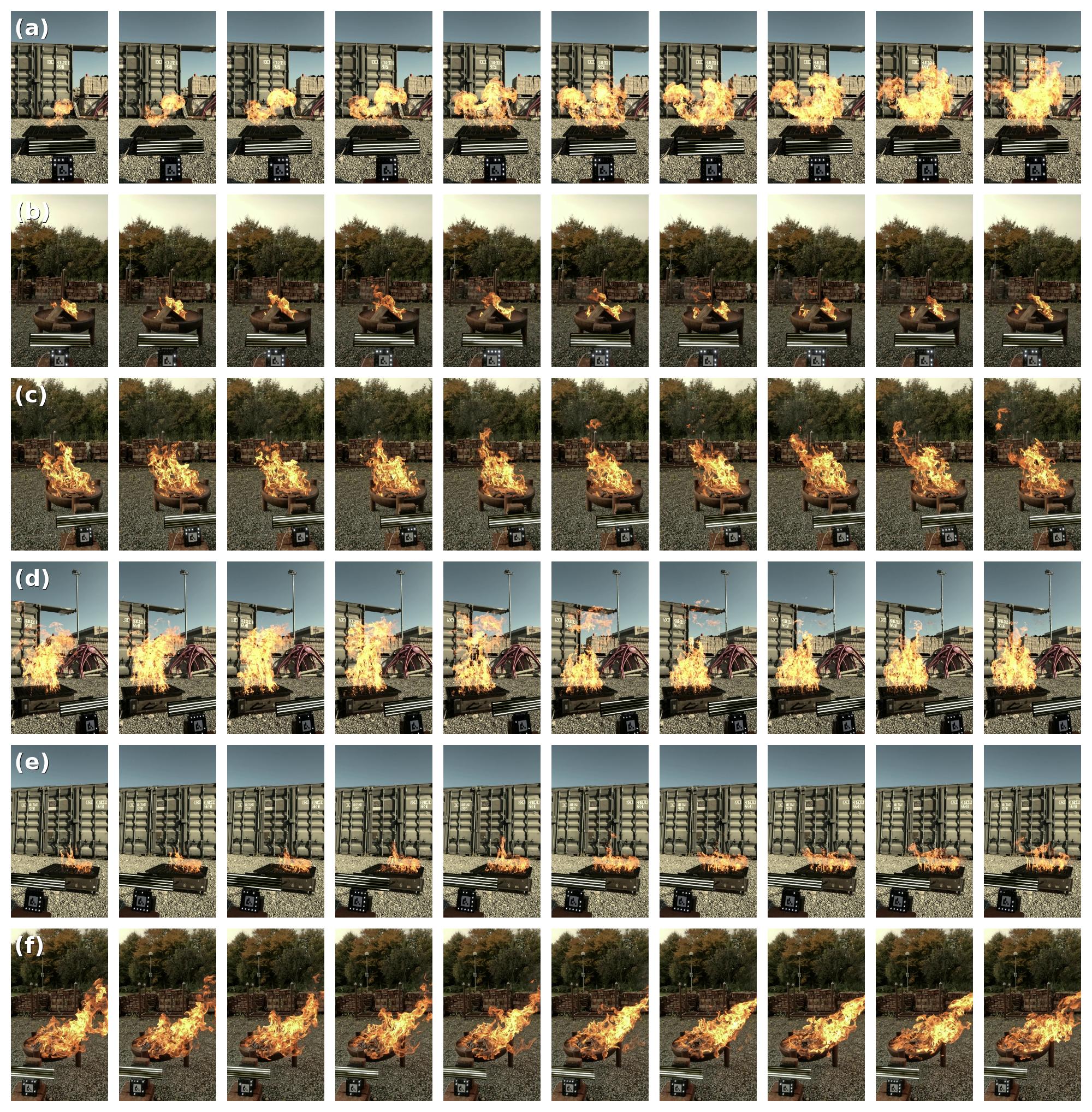}
    \vspace{-6mm}
    \caption{Qualitative overview of our real-world fire dataset. Sequences (a,d,e) depict a propane jet of varying intensity, (b) is comprised of wooden logs, (c) are sheets of cardboard, and (f) is a pile of crumpled paper. Each scene progresses temporally from left to right.
    }
    \vspace{-4mm}
    \label{fig:dataset-real}
\end{figure*}

For the validation of our method's visual reconstruction capabilities, we captured a diverse dataset of different fuels and backgrounds. We provide a more detailed overview of all captured scenes in \cref{tab:dataset}. Each video has a duration of \qty{15}{\second}, a resolution of \(1280 \times 720\), and a frame rate of \qty{400}{\hertz}. To capture the flames in their full height, all cameras were rotated by \qty{90}{\degree} to capture the scene in a portrait format. \cref{fig:setup} shows our capturing setup as well as the visual synchronization hardware while \cref{fig:dataset-real} gives an overview of the captured videos.

\begin{table*}[h!]
\centering
\caption{Summary of fire scenes, fuel types, and visual flame descriptions.}
\renewcommand{\arraystretch}{1.2}
\setlength{\tabcolsep}{6pt}
\scalebox{0.9}{
\begin{tabular}{lll}
\hline
\textbf{Scene} & \textbf{Fuel} & \textbf{Description} \\
\hline
1 & Propane & Low-intensity flame with mild wind from the left \\
2 & Propane & High-intensity flame without wind; detached upper flames \\
3 & Propane & Very low-intensity flame under quiescent conditions \\
4 & Propane & Rapidly growing flame; pronounced vortical structures \\
5 & Propane & Medium-height, wide flame exhibiting rapid growth \\
6 & Propane & Sustained high flame with continuous combustion \\
7 & Wood + Gasoline & Low-intensity flames propagating around wooden logs \\
8 & Gasoline & Medium-scale flames with visible smoke production \\
9 & Cardboard + Ethanol & Small, semi-transparent flames with visible smoke \\
10 & Cardboard + Gasoline & Large flames under wind from the right \\
11 & Cardboard & Small residual flames with visible ash formation \\
12 & Cardboard + Gasoline & Large flames under wind from the right \\
13 & Cardboard & Low-intensity residual flames with visible ashes \\
14 & Paper + Gasoline & Very large flames under wind from the left \\
15 & Paper & Small residual flames with visible ashes \\
16 & Wood + Ethanol & Medium-scale flames; wood mostly unburnt \\
17 & Wood + Ethanol & Small flames; wood partially charred \\
\hline
\end{tabular}
}
\label{tab:dataset}
\end{table*}

\subsection{Simulation}
\label{supp:dataset-simulation}
\begin{figure*}
    \centering
    \includegraphics[width=\linewidth]{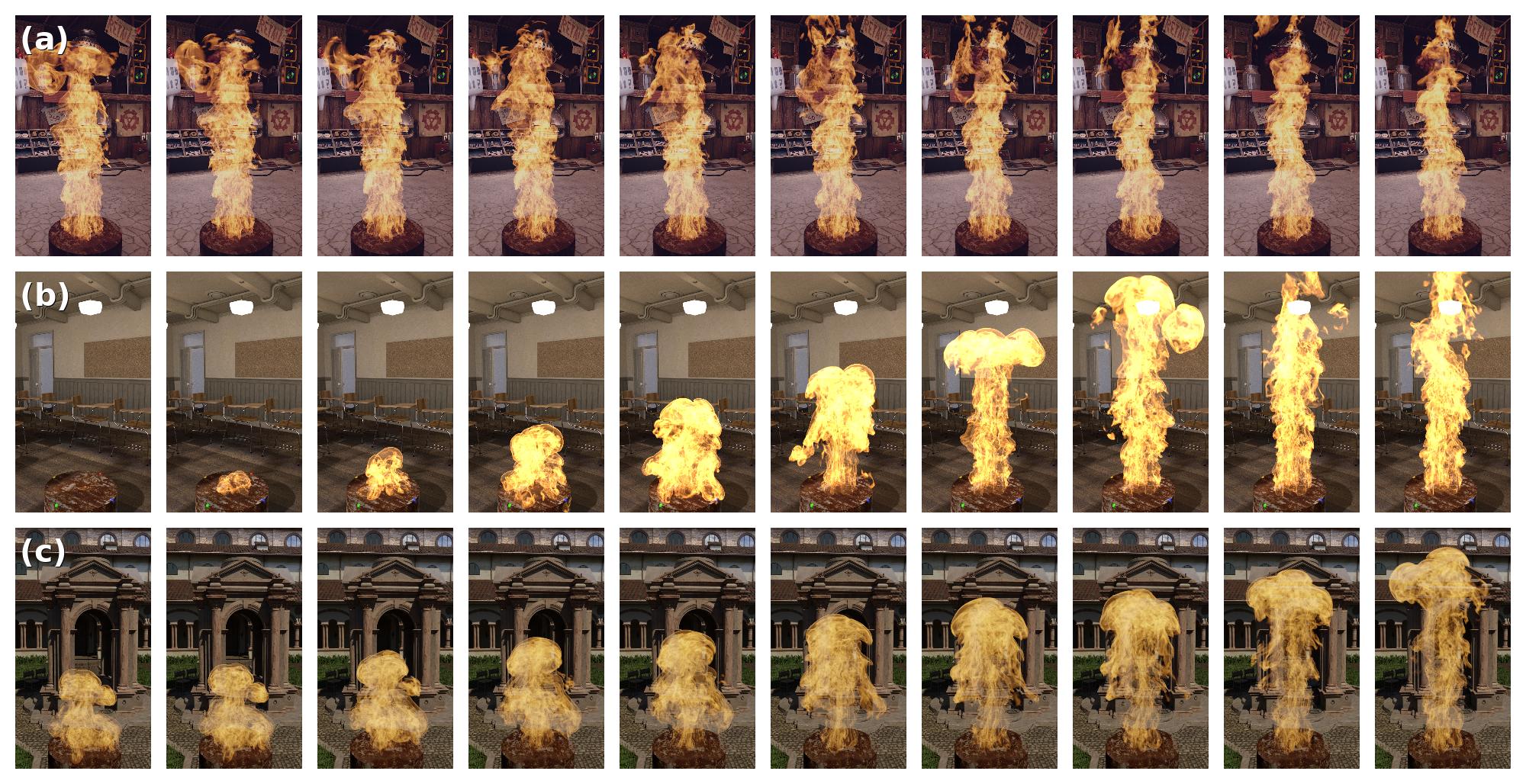}
    \vspace{-6mm}
    \caption{Qualitative overview of our synthetic fire dataset. Sequence (a) depicts the \textit{junkshop}, (b) \textit{classroom}, and (c) the \textit{monk} scene. Each scene progresses temporally from left to right.
    }
    \vspace{-4mm}
    \label{fig:dataset-sim}
\end{figure*}

For the validation of our method's dynamic motion reconstruction capabilities, we created three separate \textit{Mantaflow} fire simulations and ray-traced them in Blender.
Similarly to the real-world dataset, we captured the scene using three rotated cameras with a resolution of \(1280 \times 720\). For a feature-rich background, we chose to use the three public \textit{classroom} (Classroom, CC0, by Christophe Seux), \textit{junkshop} (The Junk Shop, CC-BY, by Alex Treviño), and \textit{monk} (Lone Monk, CC0, by Carlo Bergonzini) Blender scenes. We provide a qualitative overview of the rendered scenes in \cref{fig:dataset-sim}.

\begin{figure*}
    \centering
    \includegraphics[width=0.5\linewidth]{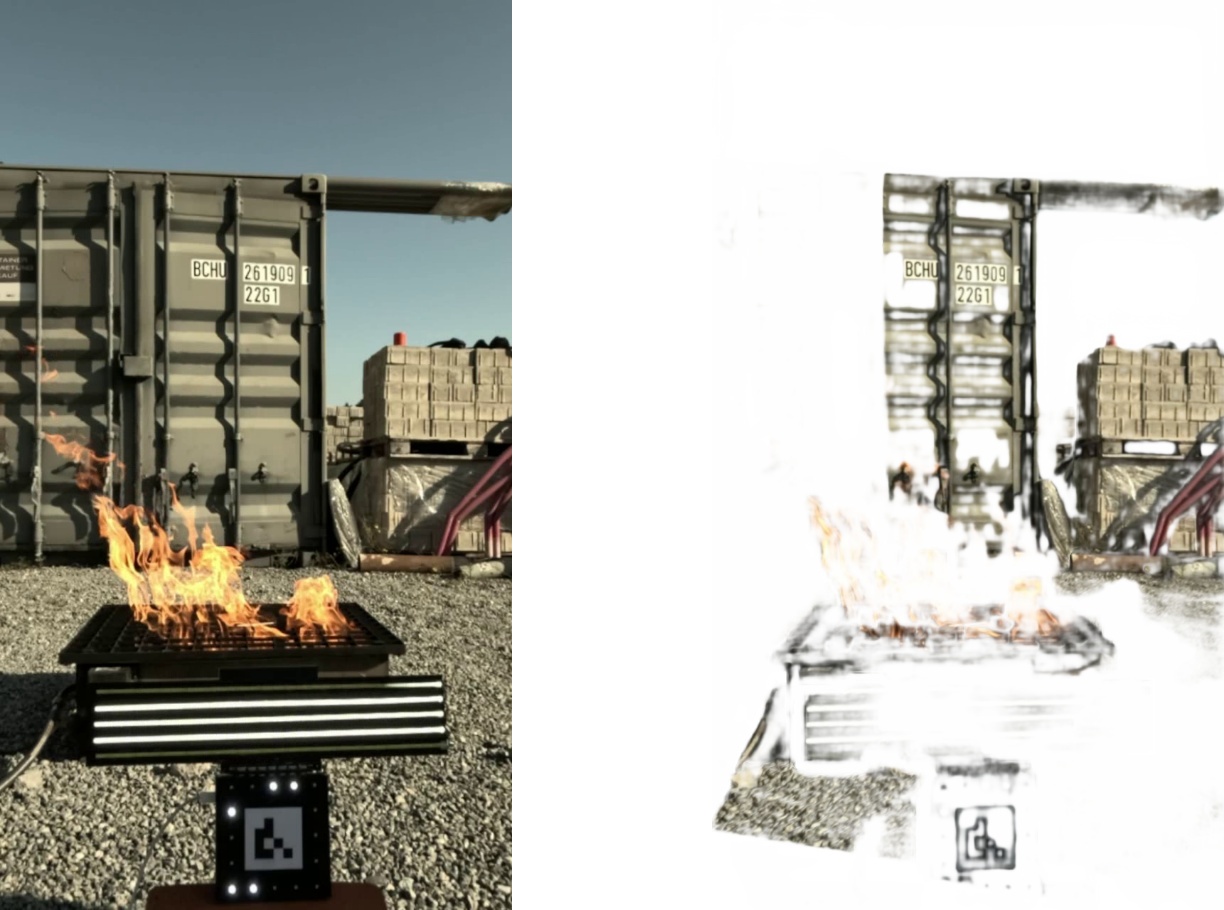}
    \vspace{0mm}
    \caption{Matched points using RoMa~\cite{edstedt2024roma} for the real-world fire scene. The left image shows the view Ifrom the middle camera and the right image shows the predicted dense matches to the left neighboring camera. Even though the fire is fully observed from both views, RoMa only successfully matches parts of the flame.
    }
    \vspace{0mm}
    \label{fig:roma}
\end{figure*}

\section{Per-Scene Results}
\label{supp:results-details}
\begin{figure*}
    \centering
    \includegraphics[width=\linewidth]{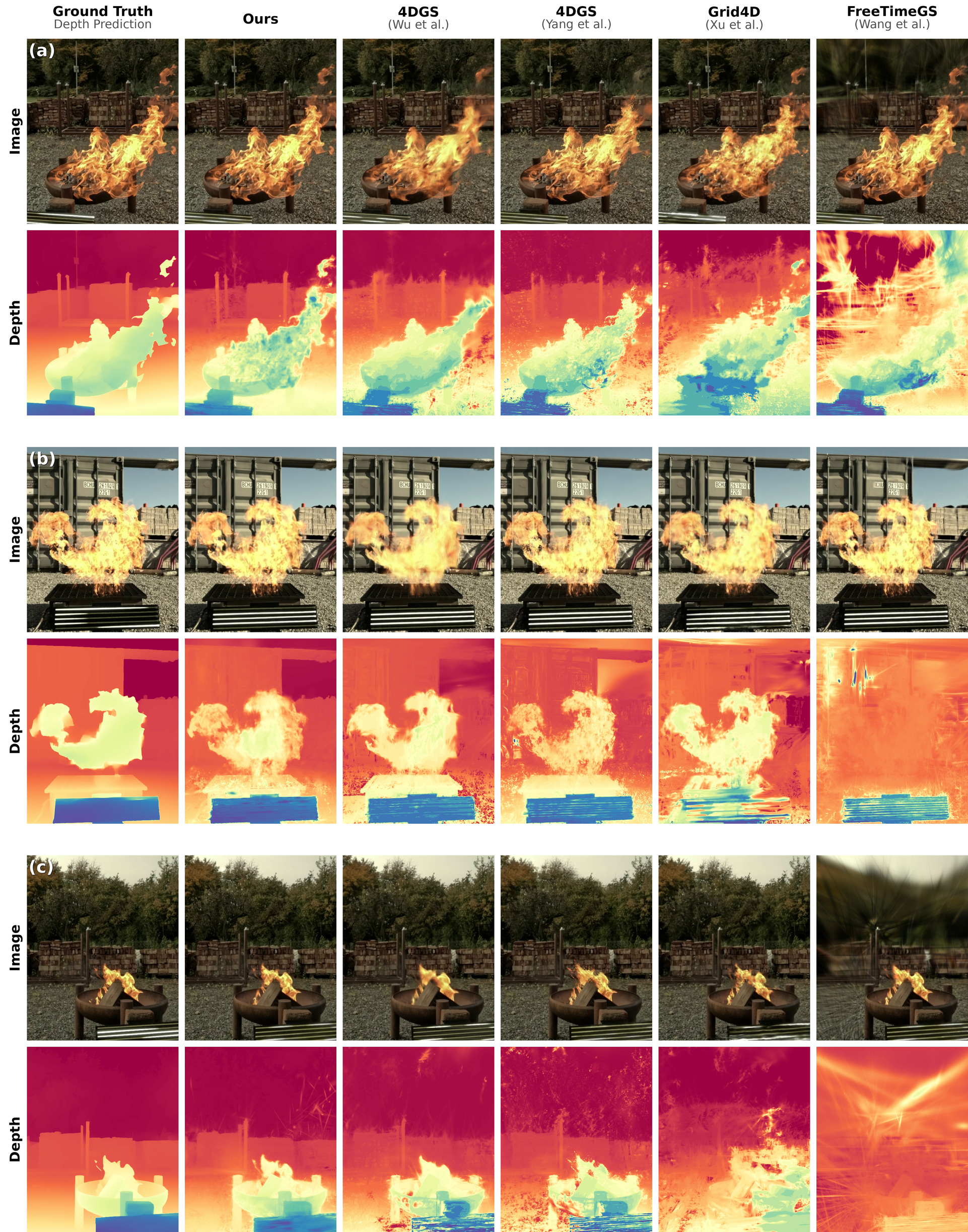}
    \vspace{0mm}
    \caption{Qualitative reconstruction results for multiple real-world scenes. Aside from our proposed method, all baselines suffer from visual or depth degradation.
    }
    \vspace{0mm}
    \label{fig:real-world-images}
\end{figure*}

In \cref{tab:per_scene_results}, we report the per-scene results of our model. The PSNR of the flame varies from \(27.24\) for scene 6 to \(34.02\) for scene 13. In \cref{fig:real-world-images}, we provide additional qualitative results for multiple scenes. We also include per-scene full-sequence videos containing the ground truth, rendered RGB, and rendered depth on the project website.

\begin{table*}
\centering
\renewcommand{\arraystretch}{1.2}
\caption{Per-scene evaluation results across all real-world scenes.}
\vspace{-2mm}
\scalebox{0.9}{
\begin{tabular}{lcccccc}
\hline
\textbf{Scene} & 
\textbf{PSNR\textsubscript{flame}}~($\uparrow$) & 
\textbf{SSIM\textsubscript{flame}}~($\uparrow$) & 
\textbf{RMSE\textsubscript{depth}}~($\downarrow$) & 
\textbf{LPIPS}~($\downarrow$) & 
\textbf{PSNR}~($\uparrow$) & 
\textbf{SSIM}~($\uparrow$) \\
\hline
1 & 27.47 & 0.863 & 0.087 & 0.027 & 36.34 & 0.980 \\
2 & 27.71 & 0.842 & 0.102 & 0.034 & 33.12 & 0.958 \\
3 & 30.50 & 0.918 & 0.097 & 0.013 & 38.55 & 0.985 \\
4 & 27.62 & 0.827 & 0.117 & 0.040 & 34.98 & 0.968 \\
5 & 29.47 & 0.864 & 0.131 & 0.025 & 36.38 & 0.976 \\
6 & 27.24 & 0.823 & 0.085 & 0.044 & 32.86 & 0.955 \\
7 & 28.80 & 0.884 & 0.079 & 0.032 & 38.90 & 0.976 \\
8 & 29.88 & 0.895 & 0.084 & 0.036 & 37.52 & 0.971 \\
9 & 33.57 & 0.936 & 0.110 & 0.033 & 39.26 & 0.975 \\
10 & 29.79 & 0.892 & 0.204 & 0.042 & 35.61 & 0.960 \\
11 & 31.90 & 0.927 & 0.072 & 0.024 & 40.18 & 0.978 \\
12 & 32.22 & 0.913 & 0.090 & 0.027 & 39.22 & 0.976 \\
13 & 34.02 & 0.949 & 0.032 & 0.024 & 39.96 & 0.980 \\
14 & 30.00 & 0.894 & 0.082 & 0.038 & 36.40 & 0.965 \\
15 & 32.73 & 0.938 & 0.061 & 0.027 & 40.17 & 0.977 \\
16 & 32.23 & 0.941 & 0.054 & 0.029 & 39.79 & 0.976 \\
17 & 33.86 & 0.942 & 0.043 & 0.022 & 38.87 & 0.978 \\
\hline
\end{tabular}
}
\label{tab:per_scene_results}
\vspace{-4mm}
\end{table*}

\section{Video Metrics}
\label{supp:video-metrics}
In our main paper, we provide per-frame metrics such as PSNR or SSIM to evaluate the visual fidelity of our reconstructed scenes for held-out test frames. Aside from per-frame metrics, there are several metrics that directly predict the perceptual differences between pairs of full videos. These provide deeper insights into temporal artifacts, like rapid changes in parts of the scene or flickering. We apply the Computer Graphics Video Quality Metric~(CGVQM)~\cite{cgvqm} as a full sequence metric to compare each rendered scene with its ground truth data. Before processing, we softly masked out the synchronization pattern to focus on the dynamic fire and its static environment. \cref{tab:cgvqm_results} provides the quantitative results for this metric. Similarly to the per-frame results, our method significantly outperforms all baseline methods, with 4DGS~\cite{yang2023gs4d} achieving second-best results.

\begin{table}[htbp]
\centering
\caption{Comparison of CGVQM scores across different methods. The scores are aggregated over the scenes. Our method significantly outperforms all other baselines.}
\label{tab:cgvqm_results}
\begin{tabular}{lc}
\hline
\textbf{Method} & \textbf{CGVQM (\(\uparrow\))} \\ \hline
Ours                                   & \(87.7\pm2.0\)  \\
4DGS~\cite{yang2023gs4d}               & \(80.2\pm5.0\)  \\
Grid4D~\cite{10.5555/3737916.3741850}  & \(75.9\pm5.3\)  \\
4DGS~\cite{Wu2024FourDGS}              & \(73.5\pm7.4\)  \\
FreeTimeGS~\cite{Wang2025FreeTimeGS}   & \(43.6\pm26.7\) \\ \hline
\end{tabular}
\vspace{-4mm}
\end{table}

\section{Spatial Holdout Validation}
\label{supp:spatial-holdout}
While our method is primarily tailored towards physically plausible geometry (DVE) and temporally generalizable visuals, we can slightly adapt it to a spatial holdout setting. This is done by reducing the Gaussians' initial lifetime and increasing the lifetime's learning rate, which enables more spatial detail for individual temporal snapshots. We evaluated this using a 5-camera setup with three cameras for training (as in our real-world experiments) and two additional intermediate cameras held out for validation. \cref{tab:spatial_holdout} shows the results for this setting. We additionally included two dynamic NeRF-based baselines, K-Planes~\cite{fridovichkeil2023kplanes} and MSTH~\cite{wang2023masked}. While all Gaussian-based methods (including our method) degrade gracefully under this setting, the NeRF-based methods show a catastrophic failure due to their lack of a robust initialization.

\begin{table}[htbp]
  \centering
  \caption{Spatial holdout evaluation on synthetic scenes.}
  \renewcommand{\arraystretch}{1.2}
  \resizebox{\linewidth}{!}{%
  \begin{tabular}{lcccccc}
    \toprule
    \textbf{Method} &
    \textbf{PSNR\textsubscript{flame}}~($\uparrow$) &
    \textbf{SSIM\textsubscript{flame}}~($\uparrow$) &
    \textbf{RMSE\textsubscript{depth}}~($\downarrow$) &
    \textbf{LPIPS}~($\downarrow$) &
    \textbf{PSNR}~($\uparrow$) &
    \textbf{SSIM}~($\uparrow$) \\
    \midrule
    Ours                     & \first{$20.54 \pm 0.43$} & \first{$0.647 \pm 0.018$} & \first{$0.190 \pm 0.029$} & \first{$0.152 \pm 0.016$} & \first{$20.89 \pm 1.60$} & \first{$0.737 \pm 0.054$} \\
    FreeTimeGS               & \second{$19.48 \pm 0.41$} & \second{$0.590 \pm 0.033$} & \third{$0.228 \pm 0.018$} & $0.346 \pm 0.047$ & $17.09 \pm 0.76$ & $0.590 \pm 0.057$ \\
    4DGS\textsubscript{Yang} & \third{$19.05 \pm 0.67$} & \third{$0.577 \pm 0.036$} & $0.286 \pm 0.028$ & \second{$0.243 \pm 0.025$} & \third{$19.08 \pm 2.23$} & \second{$0.652 \pm 0.057$} \\
    4DGS\textsubscript{Wu}   & $17.94 \pm 0.38$ & $0.529 \pm 0.031$ & \second{$0.224 \pm 0.036$} & $0.287 \pm 0.050$ & \second{$19.33 \pm 2.27$} & \third{$0.620 \pm 0.115$} \\
    Grid4D                   & $17.72 \pm 0.52$ & $0.520 \pm 0.035$ & $0.244 \pm 0.053$ & \third{$0.284 \pm 0.040$} & $18.45 \pm 1.76$ & $0.578 \pm 0.096$ \\
    MSTH                     & $10.83 \pm 5.45$ & $0.351 \pm 0.136$ & $0.250 \pm 0.046$ & $0.505 \pm 0.122$ & $12.52 \pm 1.84$ & $0.407 \pm 0.098$ \\
    K-Planes                 & $\phantom{0}8.81 \pm 1.01$ & $0.344 \pm 0.050$ & $0.289 \pm 0.024$ & $0.489 \pm 0.027$ & $12.11 \pm 0.88$ & $0.451 \pm 0.082$ \\
    \bottomrule
  \end{tabular}%
  }
  \label{tab:spatial_holdout}
\end{table}

\section{Other Domains}
\label{supp:other-domains}
\begin{figure}[htbp]
  \centering
  \setlength{\tabcolsep}{8pt}
  \newcommand{\oddom}[1]{\includegraphics[width=\linewidth]{#1}}
  \begin{tabular}{@{}m{1.1em}@{\hspace{8pt}}m{0.26\linewidth}m{0.26\linewidth}m{0.26\linewidth}@{}}
     & \multicolumn{1}{c}{\footnotesize GT} & \multicolumn{1}{c}{\footnotesize Recon} & \multicolumn{1}{c}{\footnotesize Depth} \\[8pt]
    \rotatebox{90}{\footnotesize Smoke}     & \oddom{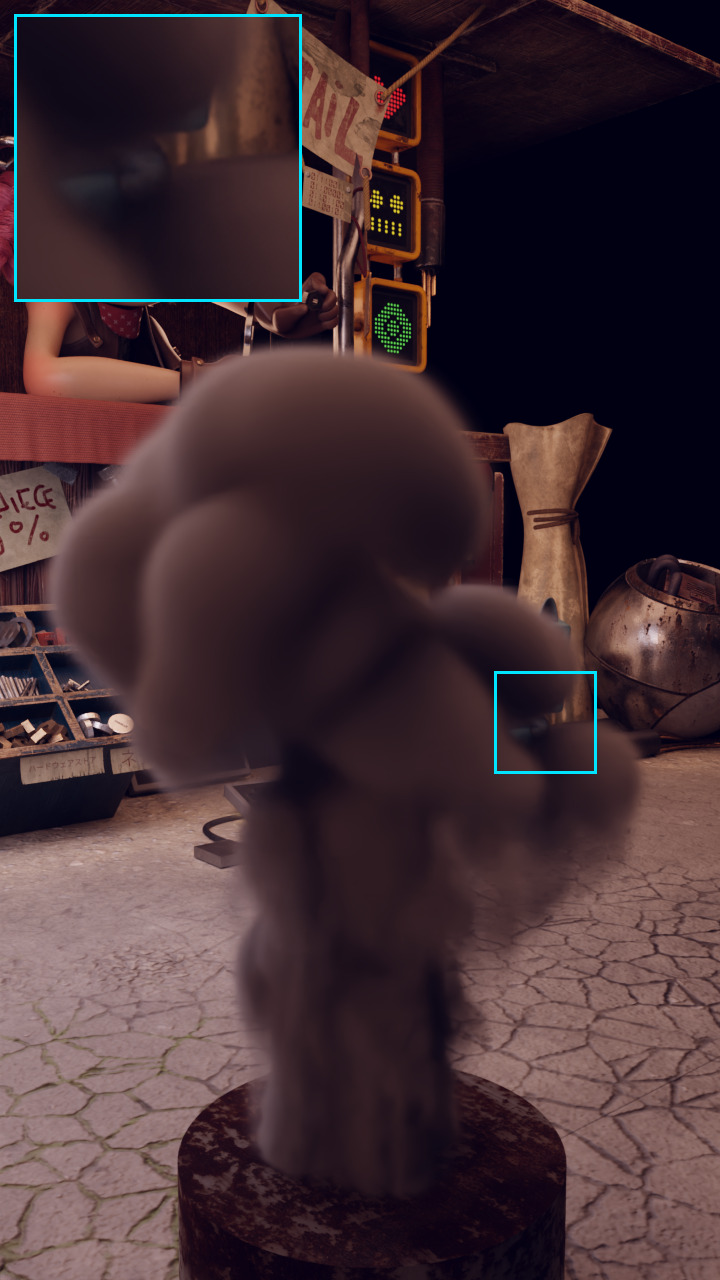}     & \oddom{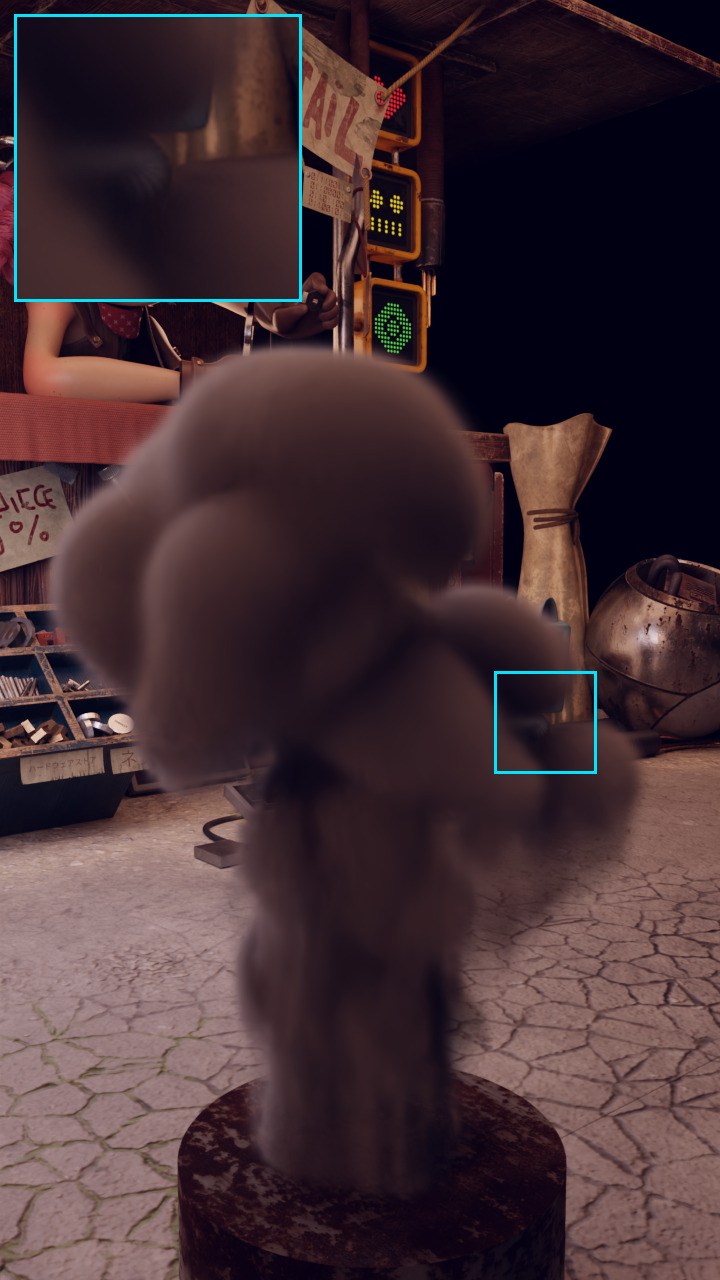}     & \oddom{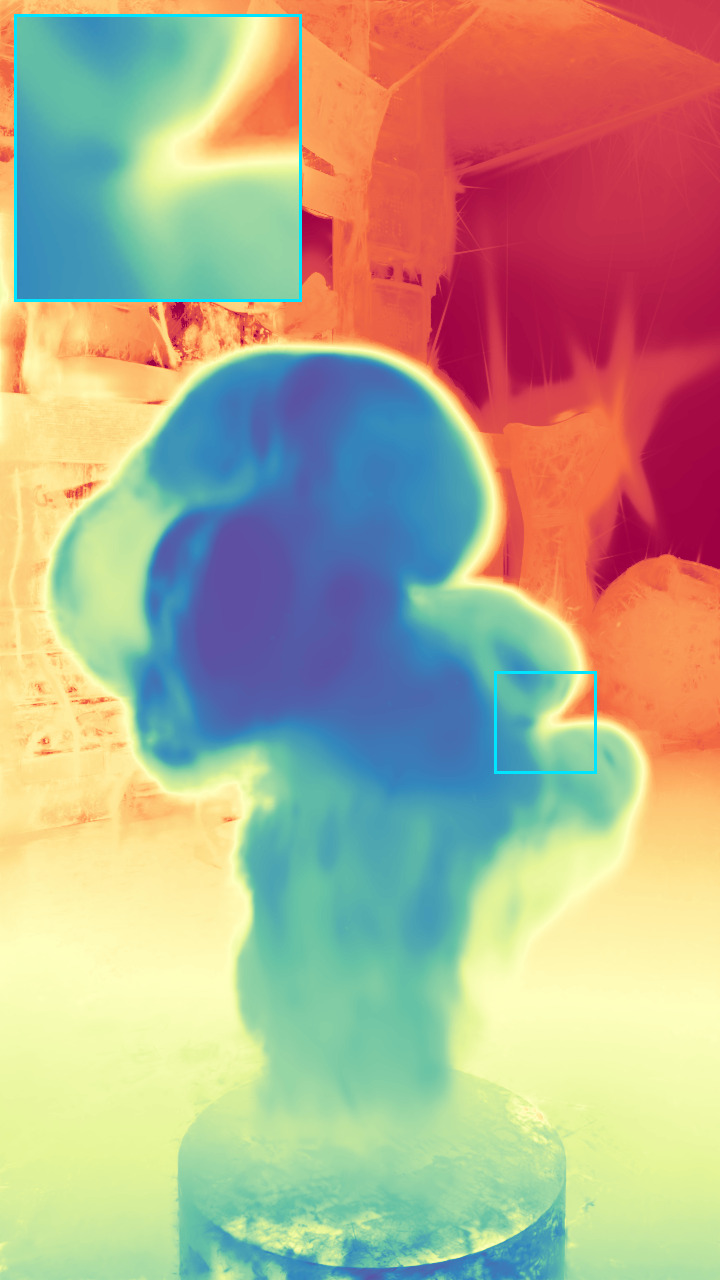} \\[8pt]
    \rotatebox{90}{\footnotesize Waterfall} & \oddom{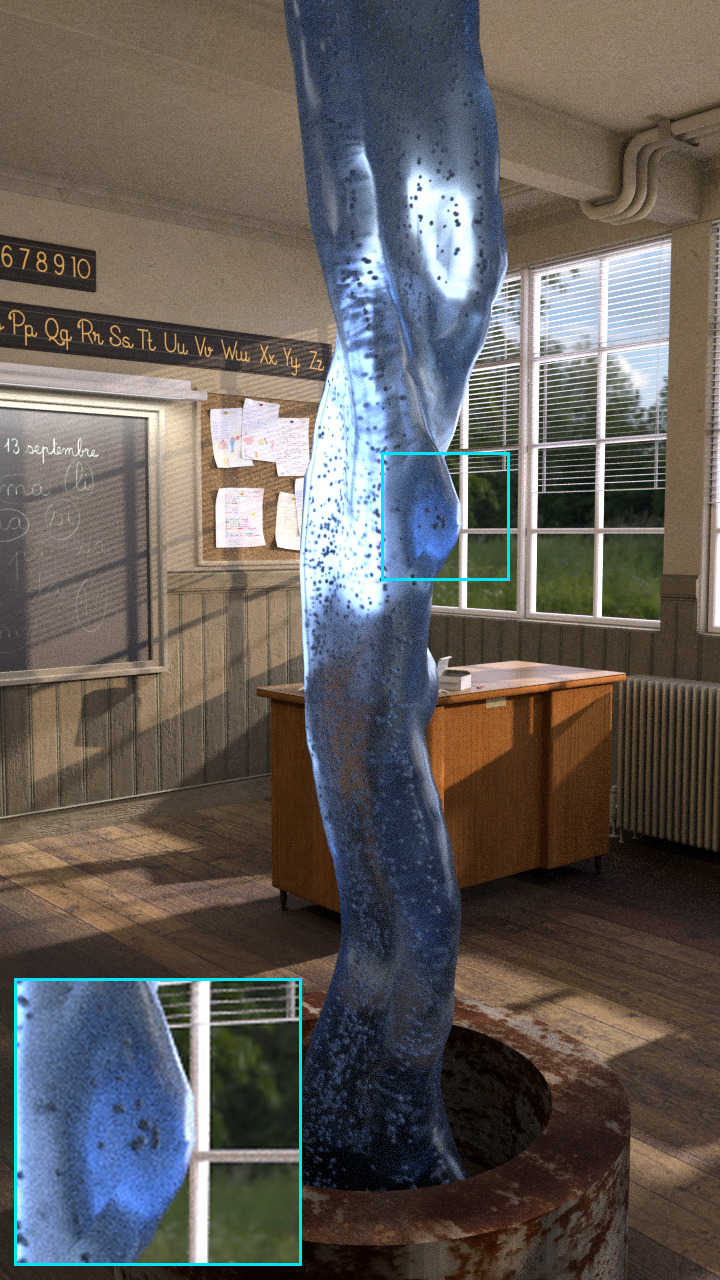} & \oddom{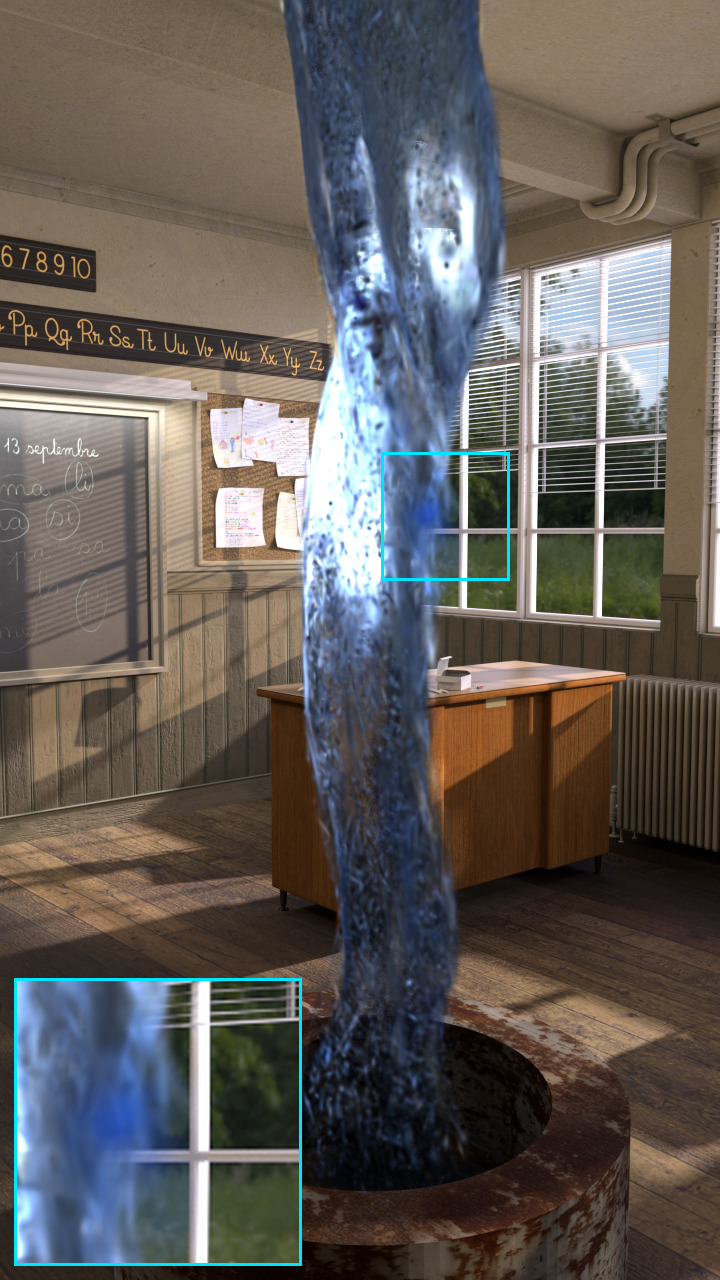} & \oddom{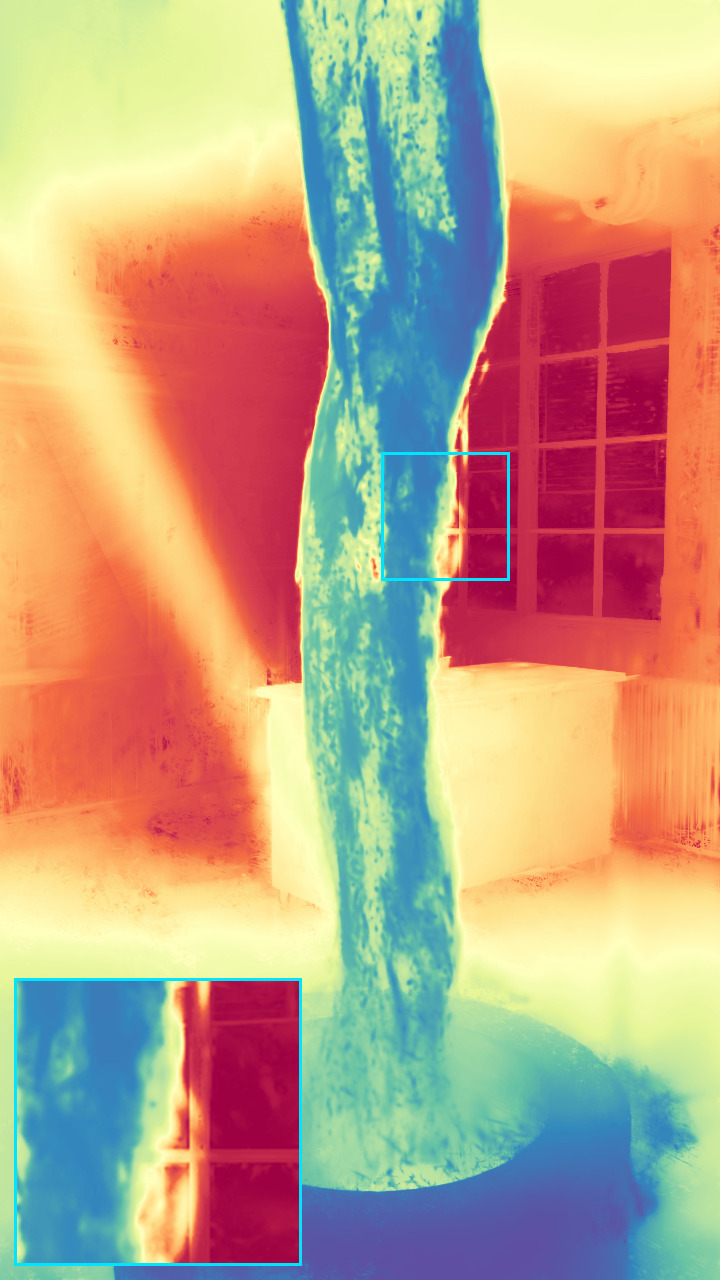} \\[8pt]
    \rotatebox{90}{\footnotesize Rain}      & \oddom{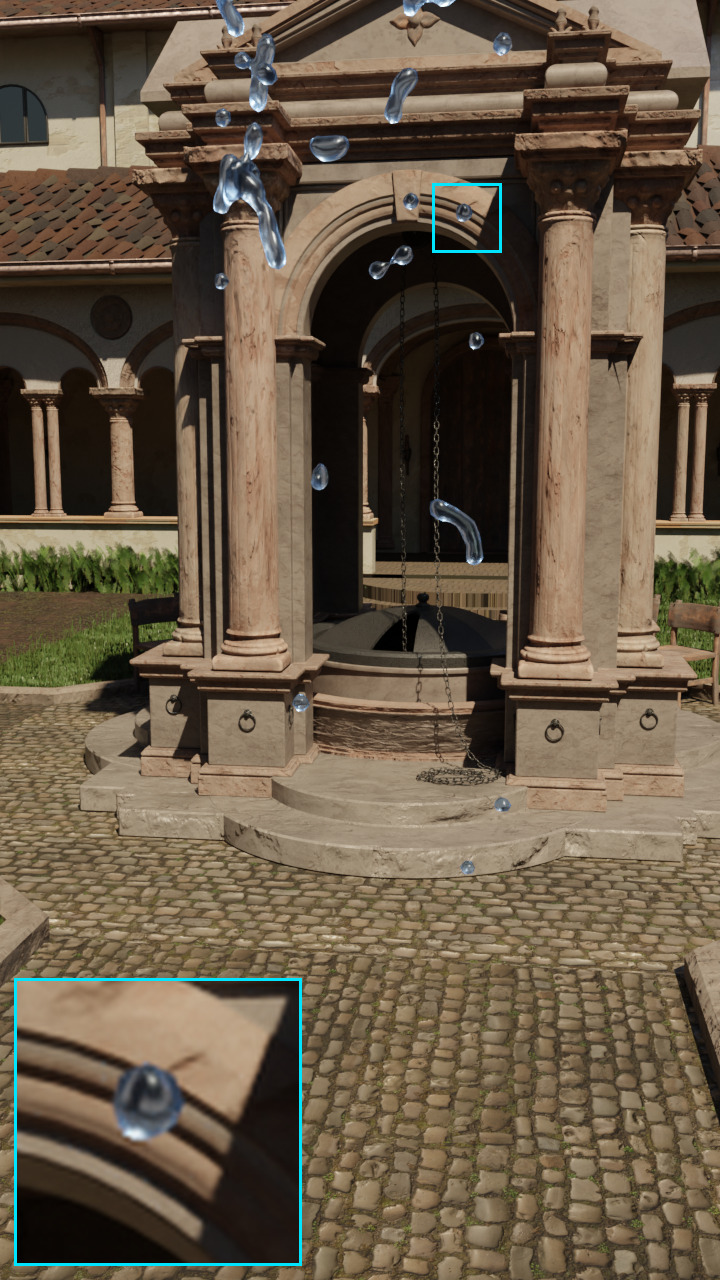}      & \oddom{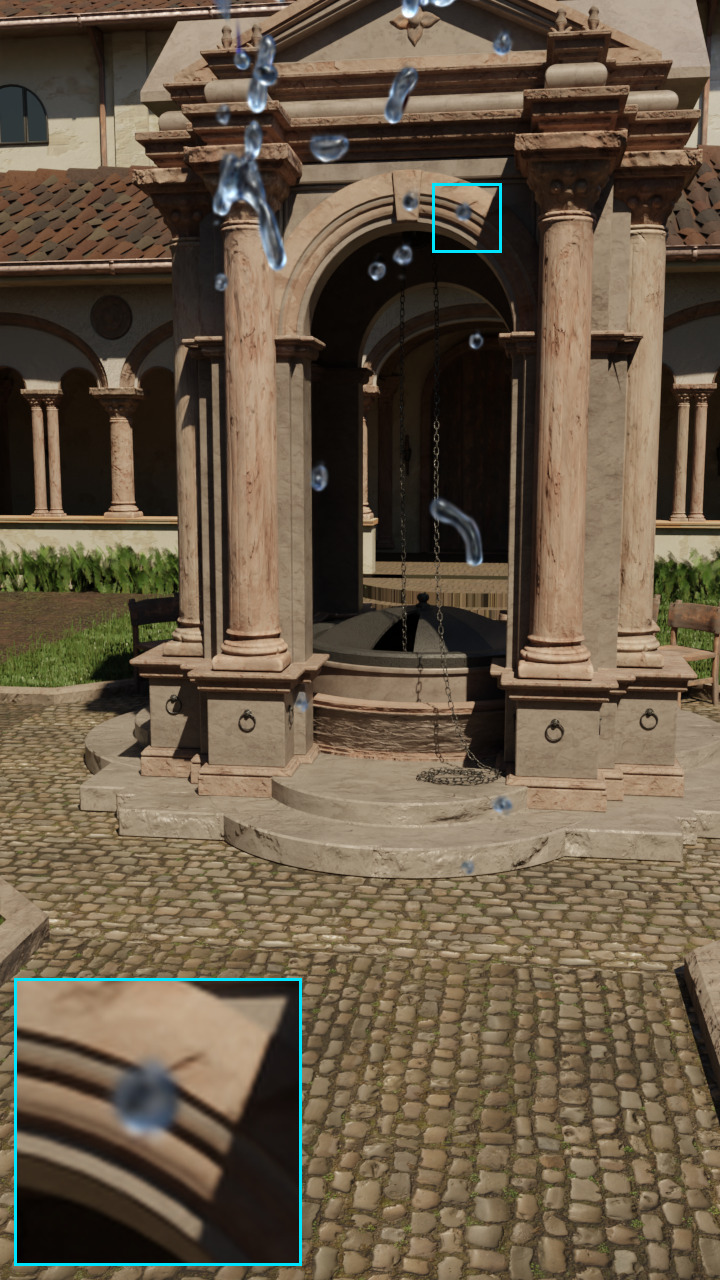}      & \oddom{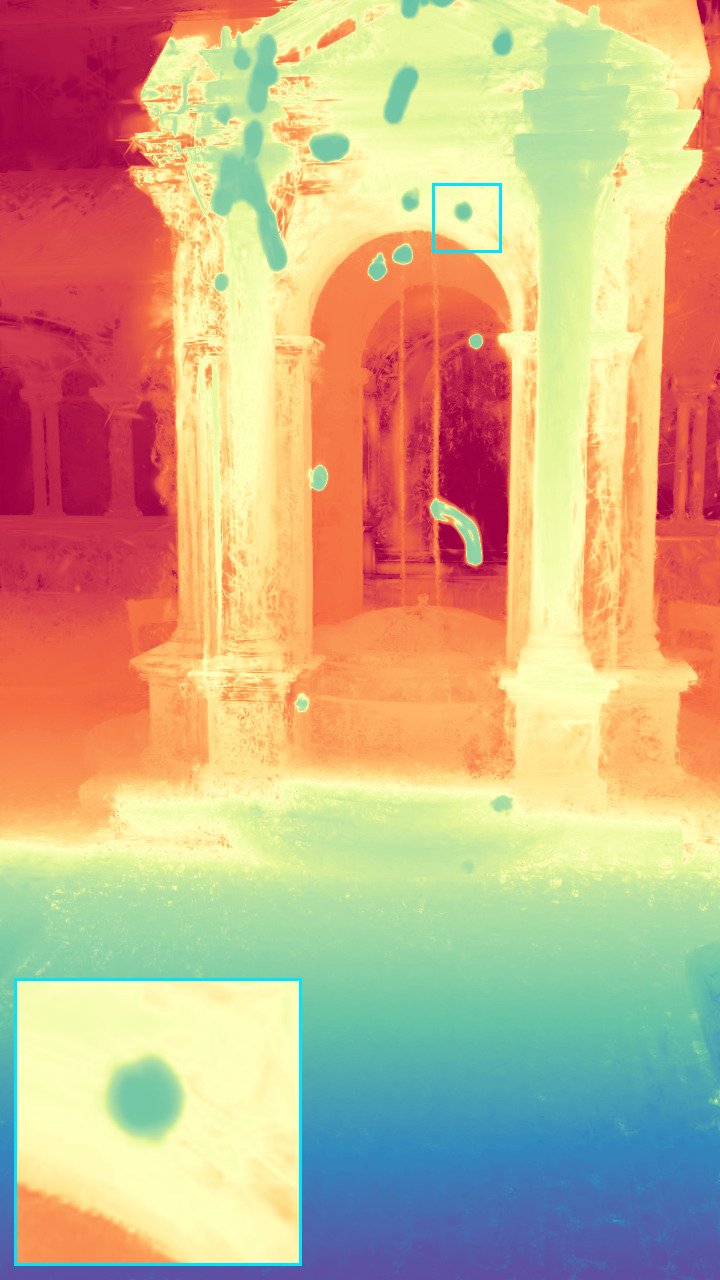}      \\
  \end{tabular}
  \caption{Qualitative results for the reconstruction of other dynamic phenomena. The first column shows the ground truth images, the middle column shows the reconstruction's rendered color while the right column shows its rendered depth.}
  \label{fig:new_domains}
\end{figure}

While our method is mainly built around reconstructing fire, the underlying pipeline is in principle also applicable to other dynamic phenomena. To test the limits of this, we rendered three additional synthetic scenes containing smoke, a waterfall, and rain. \cref{fig:new_domains} displays the qualitative results for these scenes, while \cref{tab:new_domains} holds the quantitative results. The smoke scene is much less difficult to reconstruct visually due to the lack of detailed textures, while being more difficult to reconstruct volumetrically due to the inner turbulence and the lack of strong visual cues. For the water phenomena (waterfall and rain), we notice an opposite pattern: the visual reconstruction significantly degrades due to the non-linear trajectory of the light, while the volumetric reconstruction is more accurate due to its mostly linear gravity-dominated motion. These results demonstrate that a dense flow-initialized Gaussian Splatting pipeline can generalize to other volumetric dynamic phenomena, while also highlighting the overall limitations of Gaussian Splatting for reconstructing phenomena with complex light transport or non-linear motion. We included full-sequence videos for these scenes on the project website.

\begin{table}[htbp]
  \centering
  \caption{Quantitative results for the reconstruction of other dynamic phenomena. Smoke performs best visually. The water scenes degrade visually, while being more physically accurate due to their gravity-dominated linear motion.}
  \renewcommand{\arraystretch}{1.2}
  \resizebox{\linewidth}{!}{%
  \begin{tabular}{lcccccccc}
    \toprule
    \textbf{Scene} &
    \textbf{PSNR\textsubscript{flame}}~($\uparrow$) &
    \textbf{SSIM\textsubscript{flame}}~($\uparrow$) &
    \textbf{RMSE\textsubscript{depth}}~($\downarrow$) &
    \textbf{LPIPS}~($\downarrow$) &
    \textbf{PSNR}~($\uparrow$) &
    \textbf{SSIM}~($\uparrow$) &
    \textbf{CosSim}~($\uparrow$) &
    \textbf{L\textsubscript{2}}~($\downarrow$) \\
    \midrule
    Smoke     & $38.49$ & $0.977$ & $0.124$ & $0.012$ & $43.87$ & $0.992$ & $0.733$ & $0.877$ \\
    Waterfall & $18.87$ & $0.333$ & $0.187$ & $0.197$ & $29.63$ & $0.825$ & $0.980$ & $0.510$ \\
    Rain      & $19.87$ & $0.703$ & $0.108$ & $0.014$ & $37.02$ & $0.991$ & $0.964$ & $0.295$ \\
    \bottomrule
  \end{tabular}%
  }
  \label{tab:new_domains}
\end{table}

\section{Exponential Transmittance}
\label{supp:exp-transmittance}
\begin{figure}[htbp]
  \centering
  \setlength{\tabcolsep}{4pt}
  \newcommand{\etimg}[1]{\includegraphics[width=\linewidth]{#1}}
  \begin{tabular}{@{}m{1.1em}@{\hspace{6pt}}m{0.26\linewidth}m{0.26\linewidth}m{0.26\linewidth}@{}}
     & \multicolumn{1}{c}{\footnotesize GT} & \multicolumn{1}{c}{\footnotesize Linear} & \multicolumn{1}{c}{\footnotesize Exp.} \\[3pt]
    \rotatebox{90}{\footnotesize Classroom} & \etimg{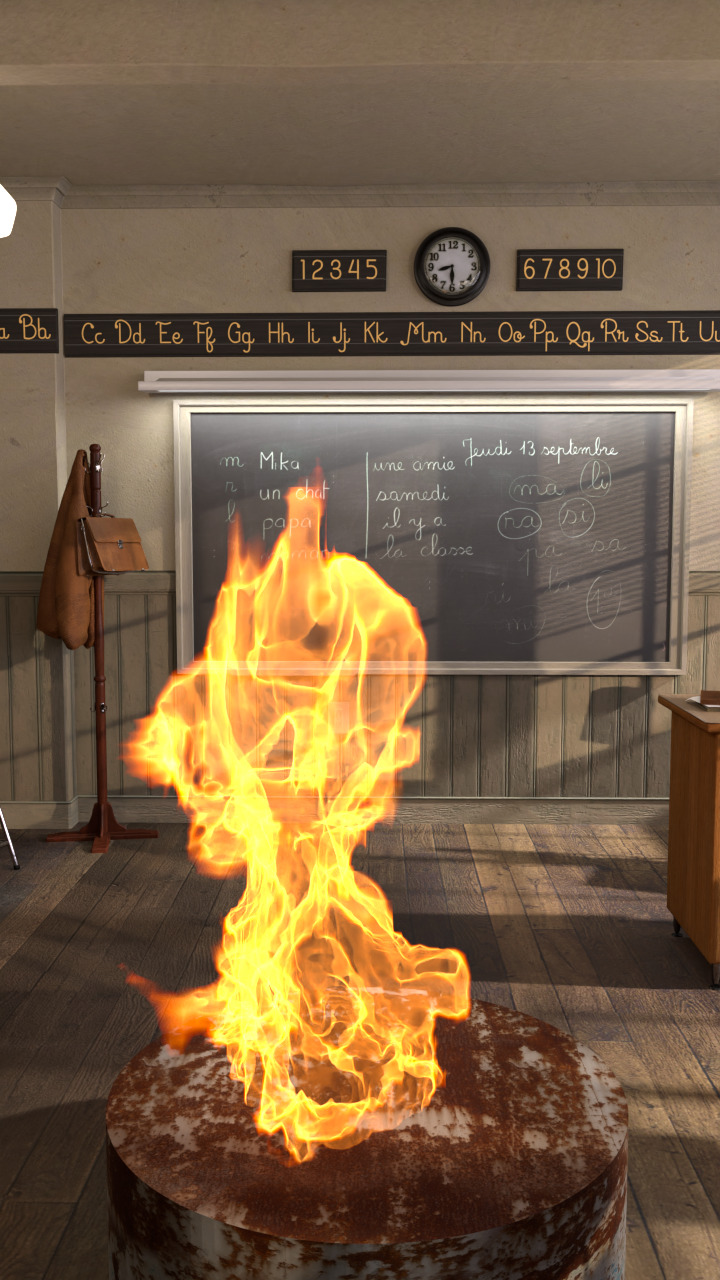} & \etimg{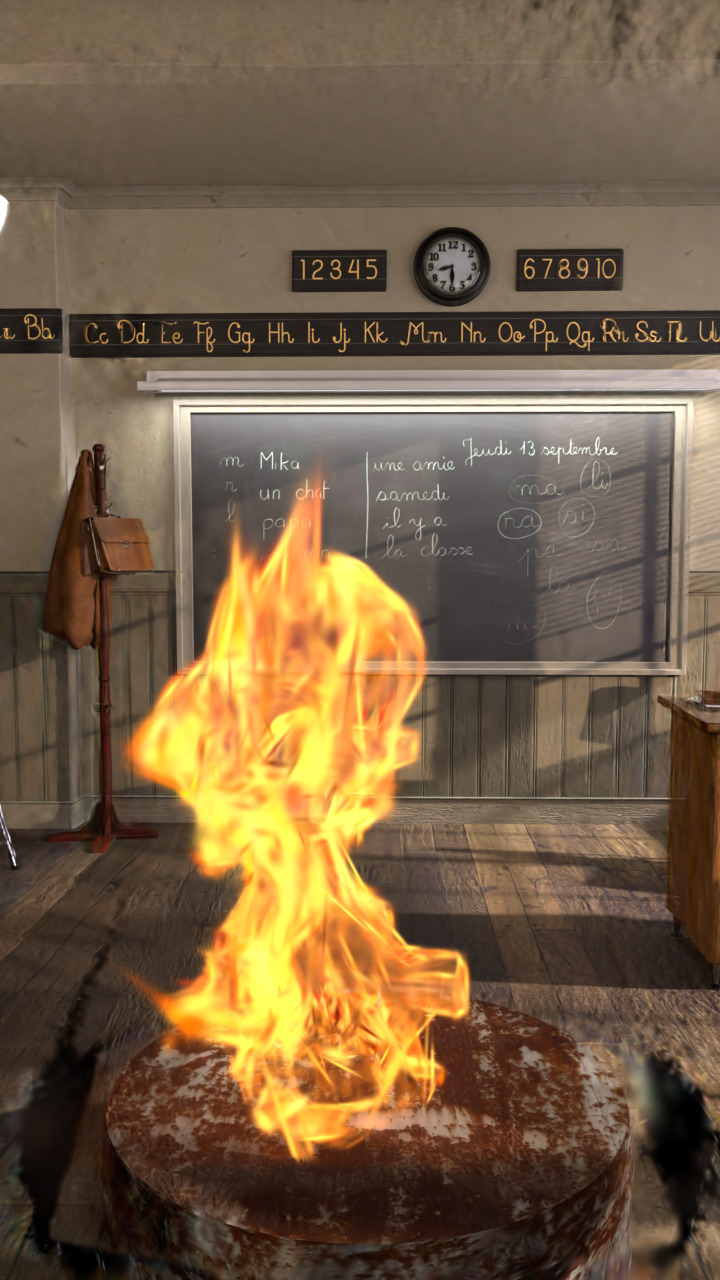} & \etimg{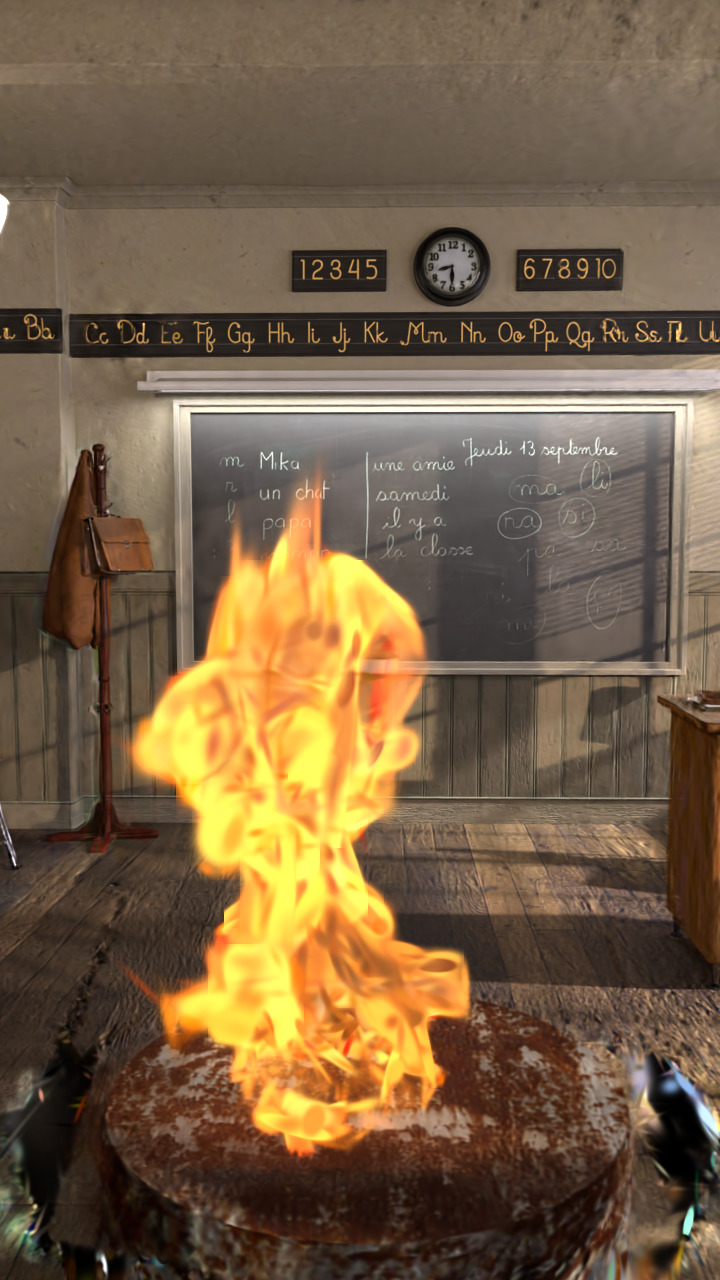} \\[3pt]
    \rotatebox{90}{\footnotesize Junkshop} & \etimg{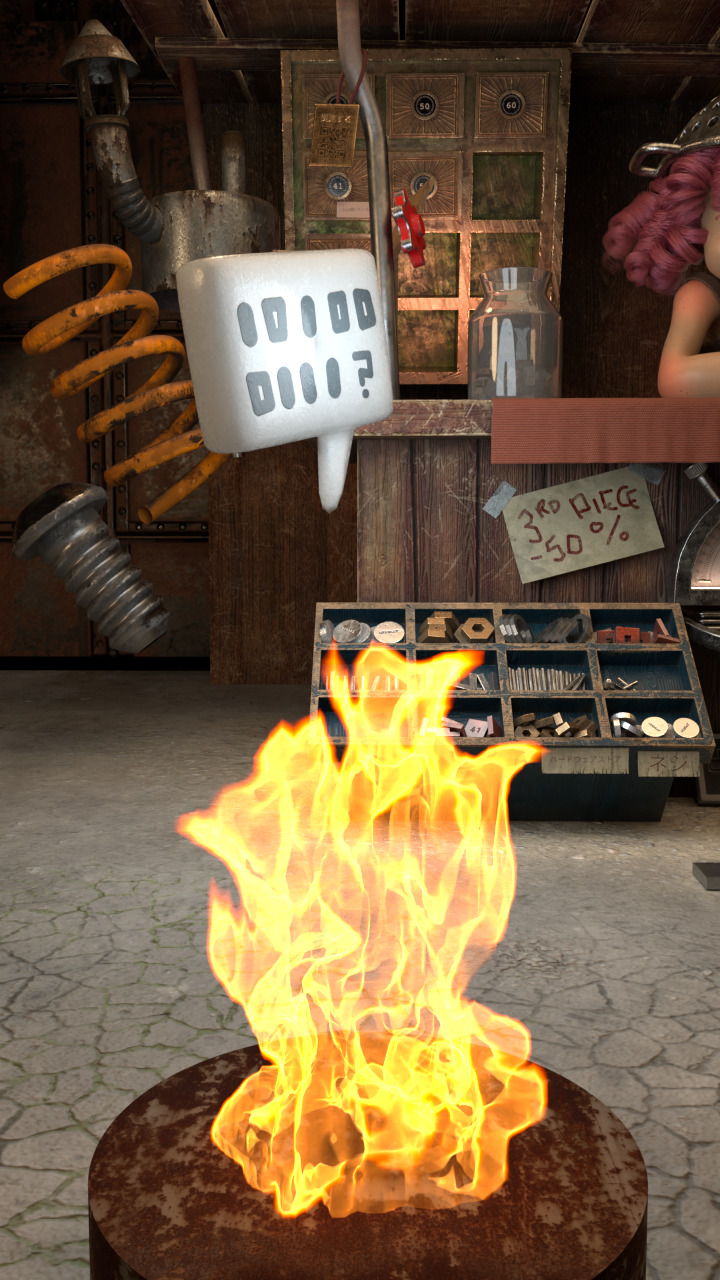} & \etimg{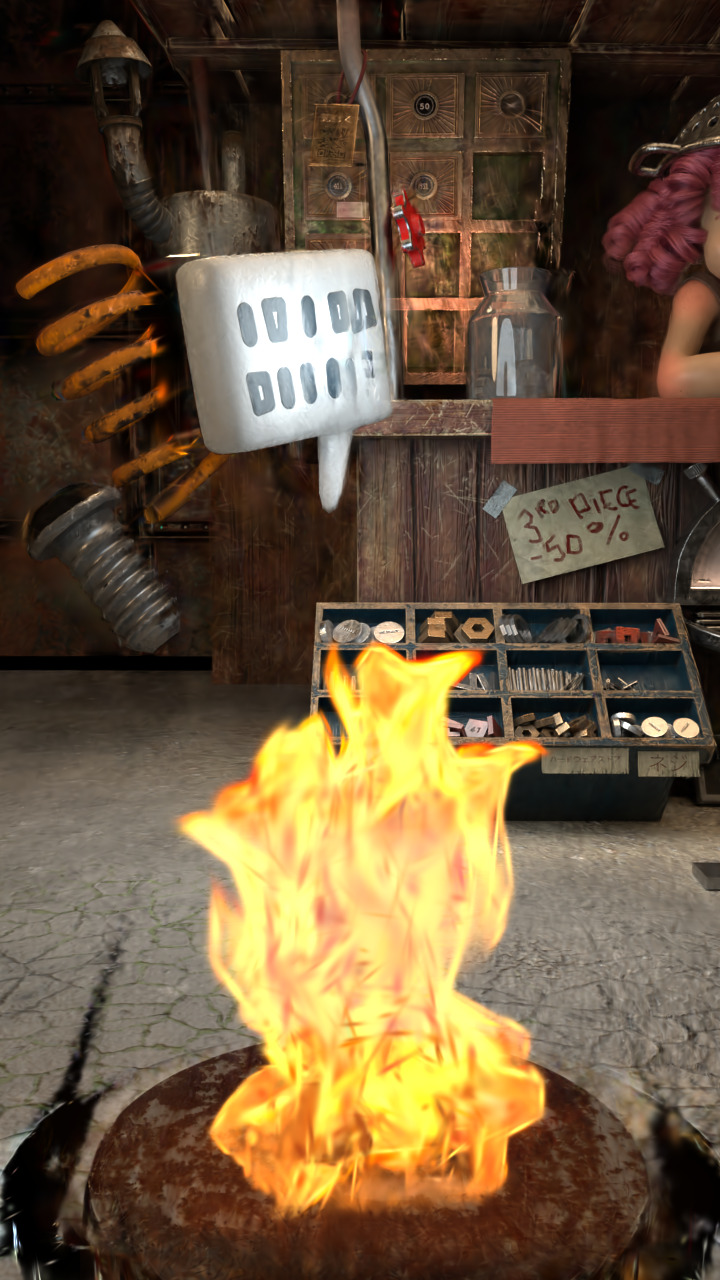} & \etimg{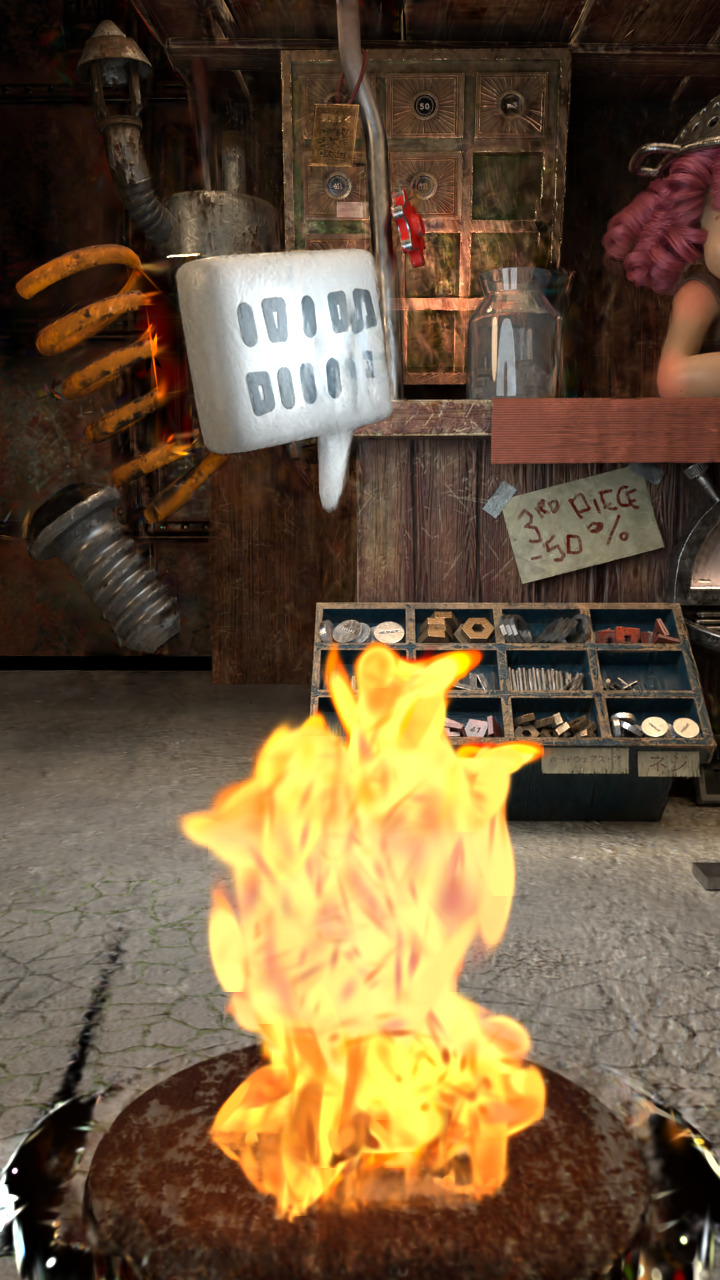} \\[3pt]
    \rotatebox{90}{\footnotesize Monk} & \etimg{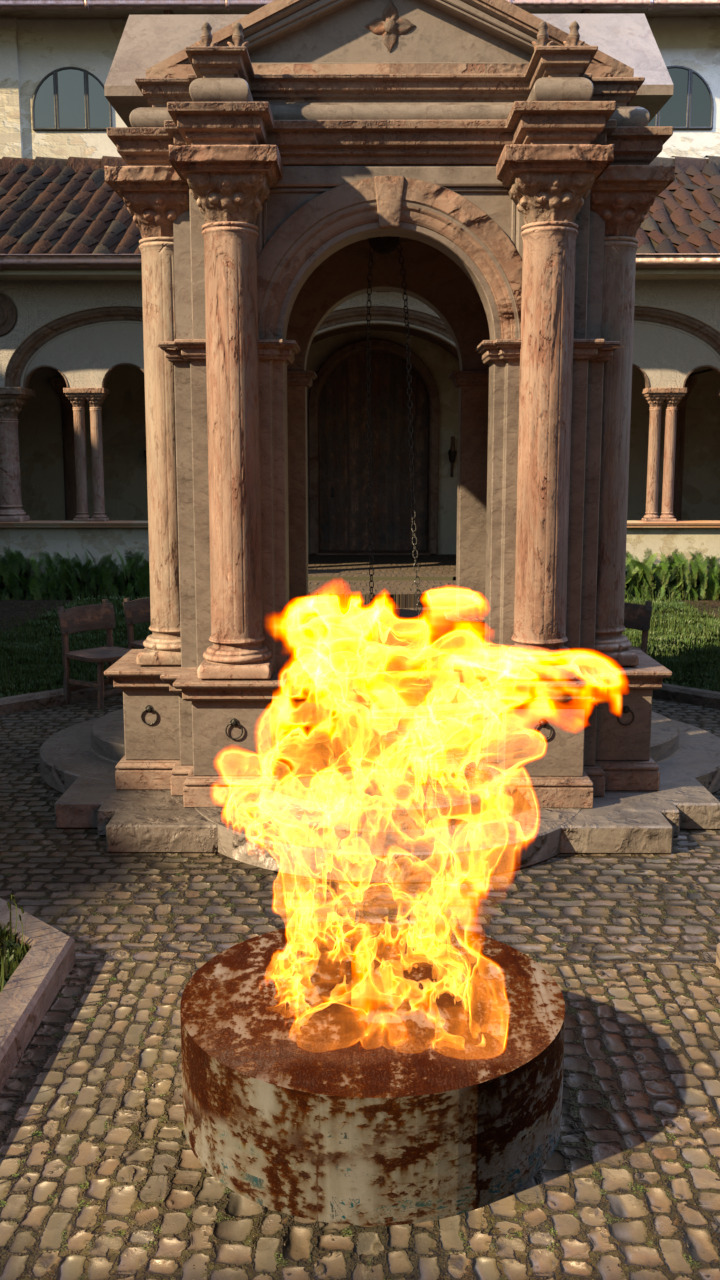} & \etimg{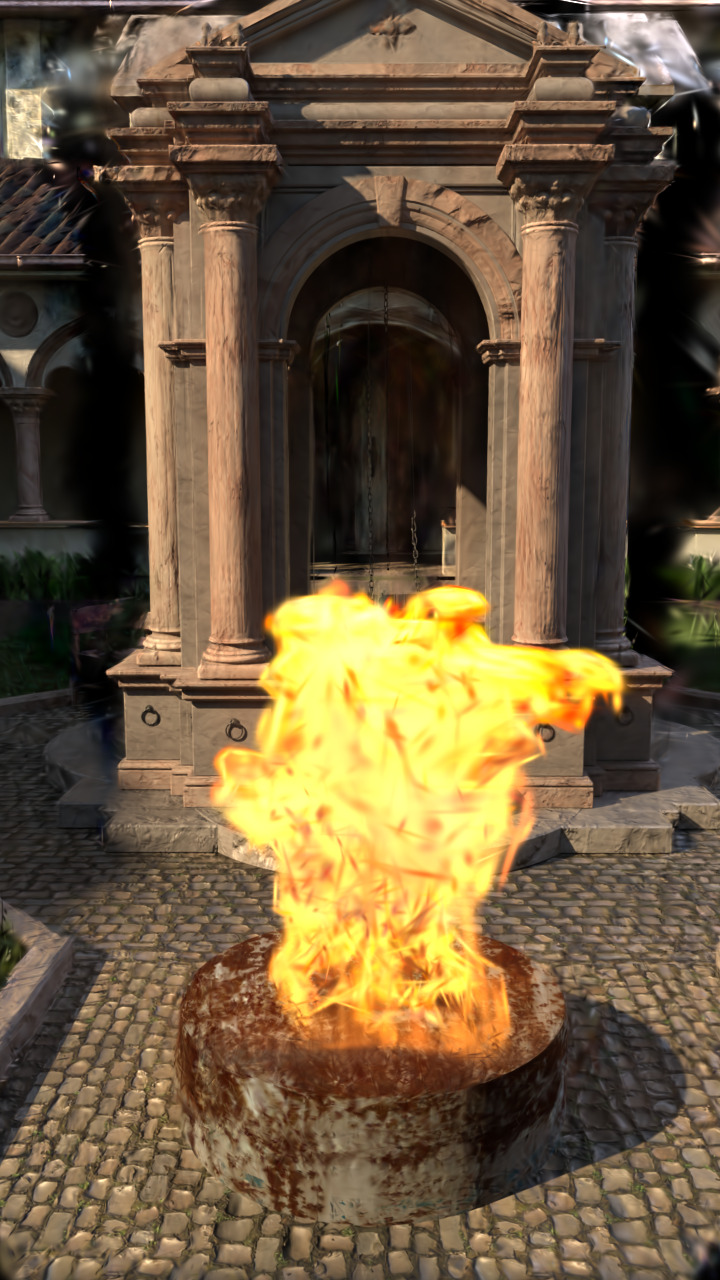} & \etimg{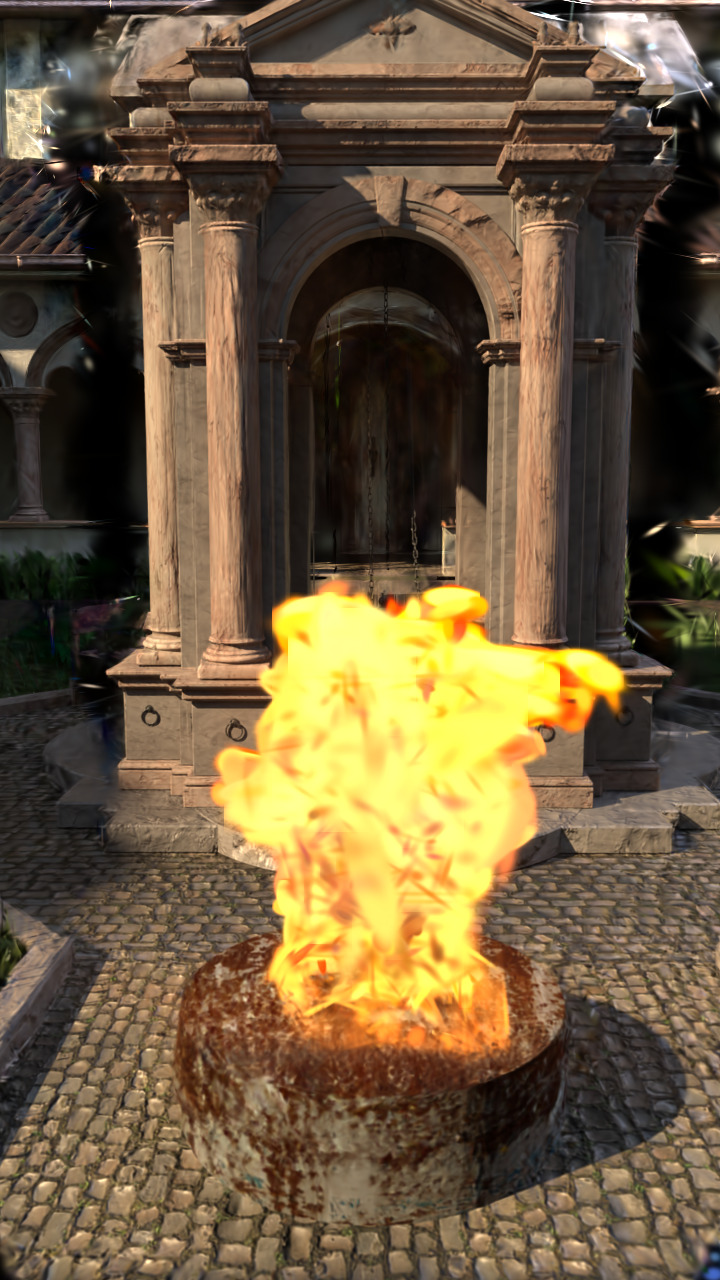} \\
  \end{tabular}
  \caption{Qualitative comparison of linear and exponential transmittance for three synthetic scenes. Aside from minor visual differences in the reconstruction artifacts, both transmittance models yield visually similar results. Depth is not included, as the reference implementation of the exponential transmittance rasterizer does not provide it as an output.}
  \label{fig:exp_transmittance}
\end{figure}

In vanilla 3D Gaussian Splatting, the transmittance is computed as a front-to-back product of per-Gaussian transmittance. While being very efficient, this is only an approximation of Beer-Lambert transmittance, which models the physically accurate exponential attenuation of light within a medium. For our scenes, this linear approximation is highly accurate due to the properties of the flame Gaussians. In our scenes, we measured each ray passing through the fire to intersect \(\approx 300\) Gaussians on average. Further, these Gaussians have an average size of half the voxel size used at initialization. For a large number of small Gaussians, the linear transmittance approaches a faithful exponential transmittance. To experimentally validate this, we reconstructed three synthetic scenes for a single point in time, once with the vanilla linear transmittance and once with an official exponential transmittance rasterizer~\cite{talegaonkar2025volumetrically} and validated them using spatially held out views. The results in \cref{tab:exp_transmittance} and \cref{fig:exp_transmittance} show no significant difference between the two rasterizers, confirming our hypothesis.

\begin{table}[htbp]
  \centering
  \caption{Quantitative comparison of linear and exponential transmittance for three synthetic scenes. While the exponential transmittance model yields slightly degraded results, the differences are minor and do not significantly affect the visual reconstruction quality.}
  \renewcommand{\arraystretch}{1.2}
  \resizebox{\linewidth}{!}{%
    \begin{tabular}{lccccc}
    \toprule
    \textbf{Transm.} &
    \textbf{PSNR\textsubscript{flame}}~($\uparrow$) &
    \textbf{SSIM\textsubscript{flame}}~($\uparrow$) &
    \textbf{PSNR}~($\uparrow$) &
    \textbf{SSIM}~($\uparrow$) &
    \textbf{LPIPS}~($\downarrow$) \\
    \midrule
    Linear & $18.51 \pm 0.56$ & $0.553 \pm 0.032$ & $20.46 \pm 1.49$ & $0.722 \pm 0.055$ & $0.170 \pm 0.016$ \\
    Exp.   & $18.16 \pm 0.56$ & $0.545 \pm 0.034$ & $20.20 \pm 1.36$ & $0.708 \pm 0.053$ & $0.197 \pm 0.019$ \\
    \bottomrule
  \end{tabular}%
  }
  \label{tab:exp_transmittance}
\end{table}

\section{Failure Cases}
\label{supp:failure-cases}
\begin{figure}[htbp]
  \centering
  \newcommand{\failurepanel}[2]{%
    \begin{minipage}[t]{0.235\linewidth}%
      \centering
      {\scriptsize\bfseries #2}\\[0.5mm]
      \includegraphics[width=\linewidth]{#1}%
    \end{minipage}%
  }
  \failurepanel{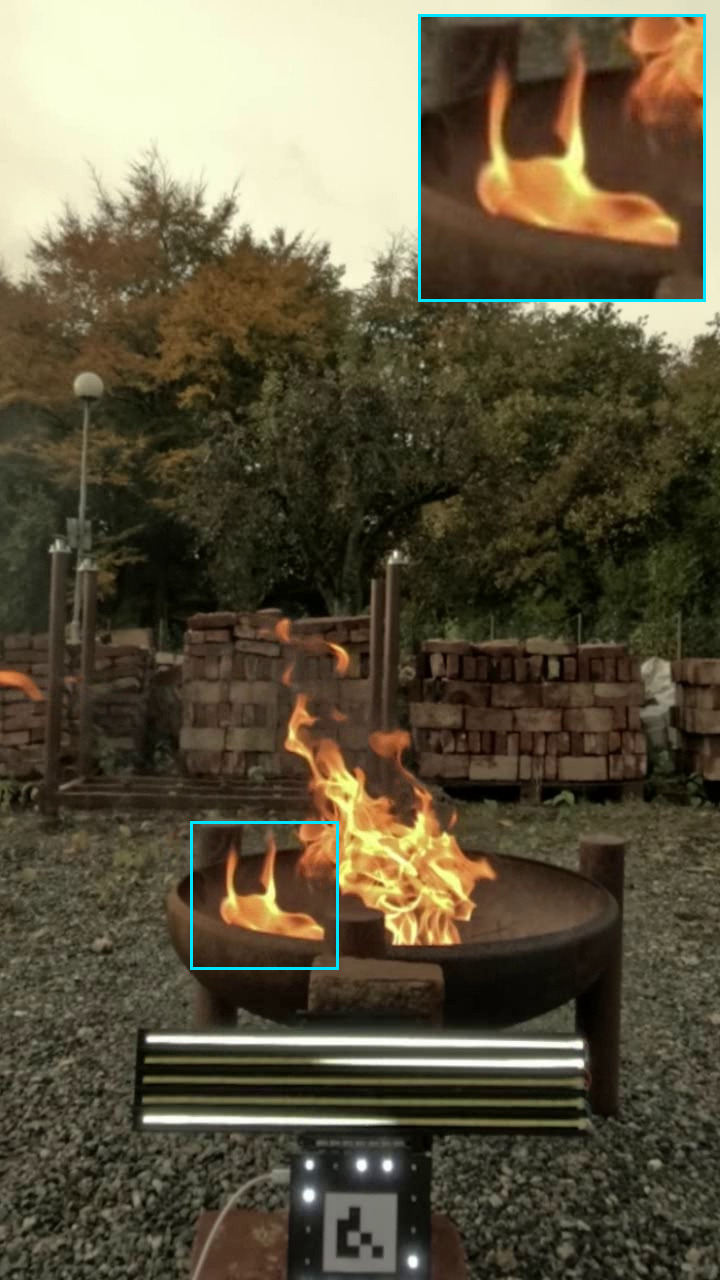}{GT \#1}\hfill
  \failurepanel{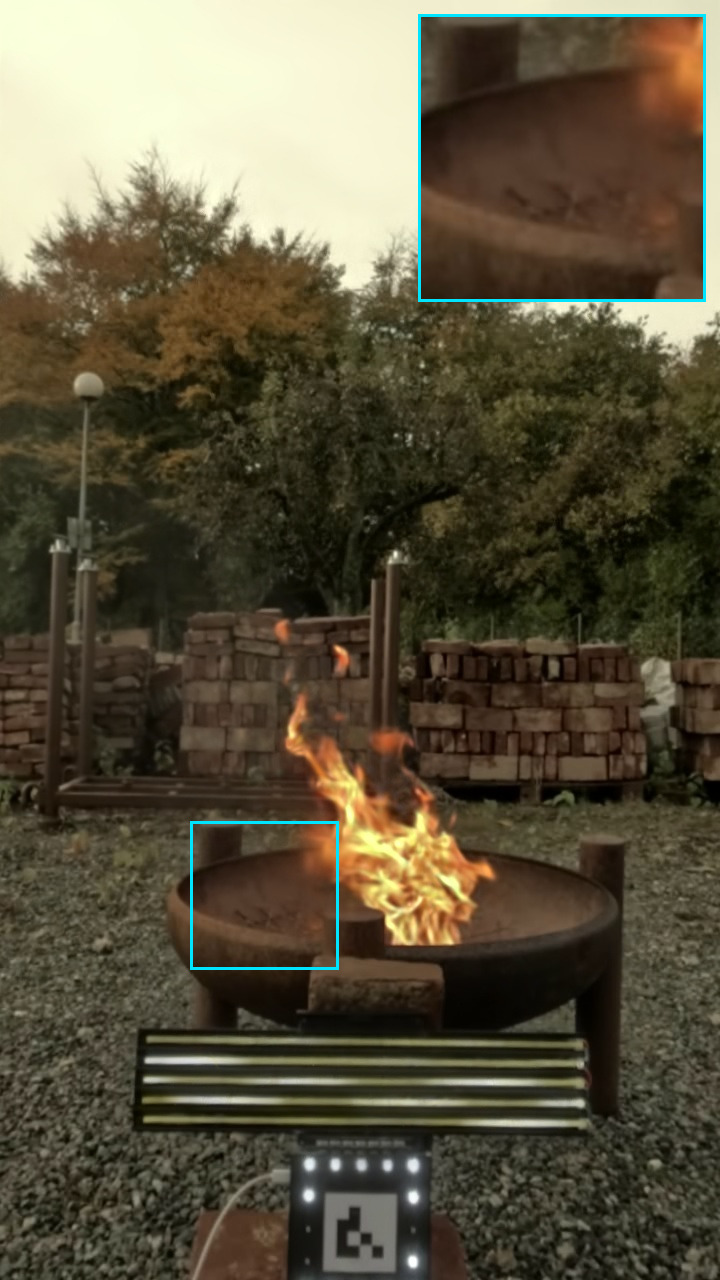}{Ours \#1}\hfill
  \failurepanel{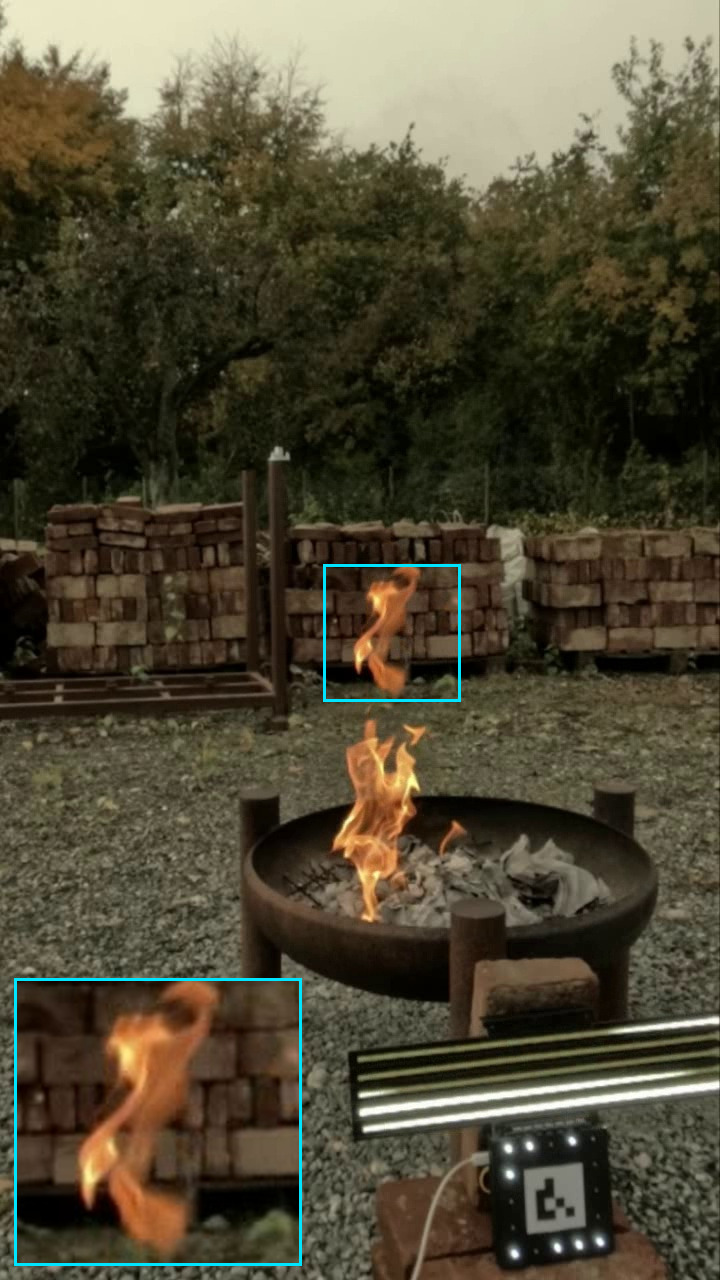}{GT \#2}\hfill
  \failurepanel{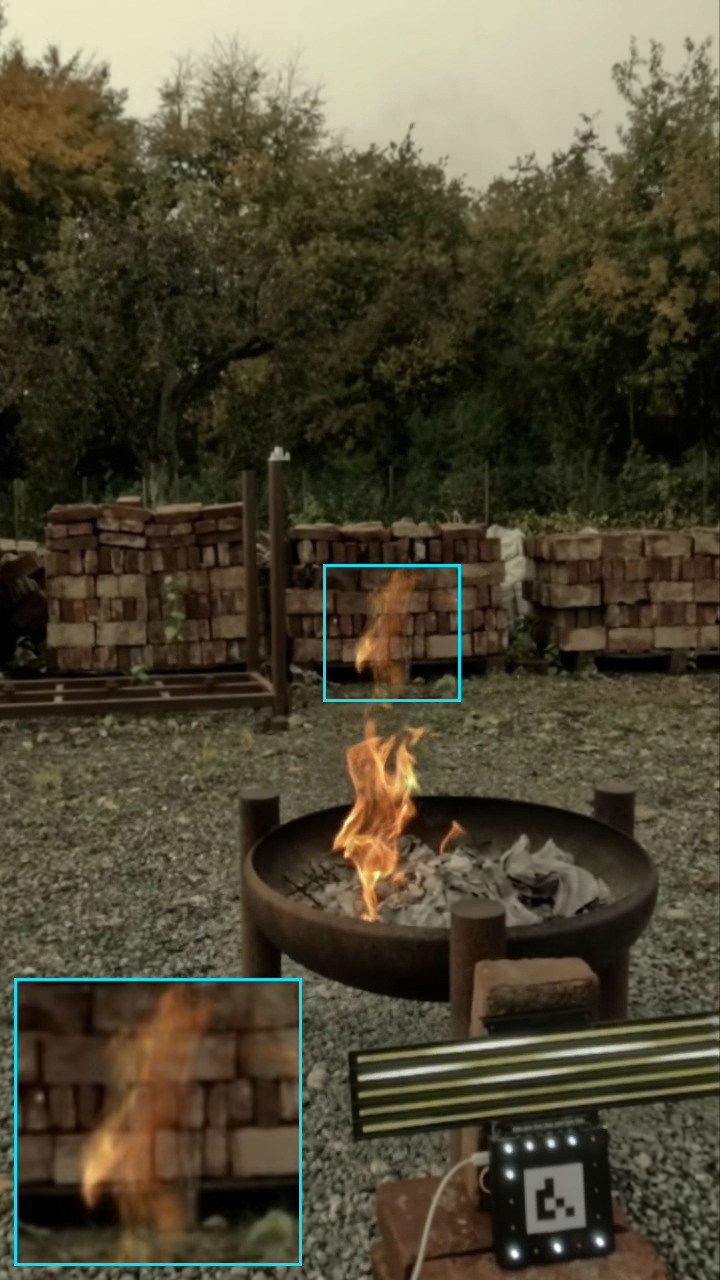}{Ours \#2}
  \caption{Failure cases for two real-world scenes.}
  \label{fig:failure_cases}
\end{figure}

Aside from the typical failure modes of Gaussian Splatting methods such as near-camera floaters (see Fig.~\ref{fig:compositions} in the main paper), we observed two main failure modes that are specific to our proposed method. Both are shown in \cref{fig:failure_cases}. The first failure mode is an incorrectly predicted 2D flow at initialization, omitting parts of the fire. These regions are not properly initialized with dynamic Gaussians, which leads to the absence of flames in the subsequent reconstruction. This optimization mostly cannot recover from this failure. The second failure mode is a blurred flame, which is a direct result of the linear motion assumption used in our whole pipeline. If this assumption is violated, the linear-motion approximation of non-linear motion leads to a smeared reconstruction. To observe these and other failure modes of our method, the full-sequence videos on the project website contain per-frame error maps for the rendered color.

\section{Novel View Trajectories}
\label{supp:novel-views}
To provide a qualitative demonstration of our model's capability to generalize to novel views, we provide a video and an interactive demo on our project website. In the video, we show the rendered results for different kinds of trajectories, leaving the camera poses used for training. For each timestep, we provide the rendered image as well as the rendered depth. To further visualize the dynamics of the reconstructed fire, we additionally render only the dynamic fire Gaussians at a reduced scale. This provides a view of the underlying motion field. For the comparison methods~\cite{Wu2024FourDGS,yang2023gs4d,10.5555/3737916.3741850,Wang2025FreeTimeGS}, we show their results along the same trajectories. All comparison methods yield significantly inferior reconstruction results, with blurry flames~\cite{Wu2024FourDGS}, flickering artifacts~\cite{yang2023gs4d,10.5555/3737916.3741850}, or deteriorated depth~\cite{Wang2025FreeTimeGS}. When focusing on the reconstructed motion field, it is clearly visible that while our method produces dense and plausible motion, all other baselines yield implausible Gaussian motion, with fire particles moving downwards or remaining stationary. 

\begin{figure}[htbp]
  \centering
  \newcommand{\rotpanel}[2]{%
    \begin{minipage}[t]{0.12\linewidth}%
      \centering
      {\scriptsize #2}\\[0.5mm]
      \includegraphics[width=\linewidth]{#1}%
    \end{minipage}%
  }
  \rotpanel{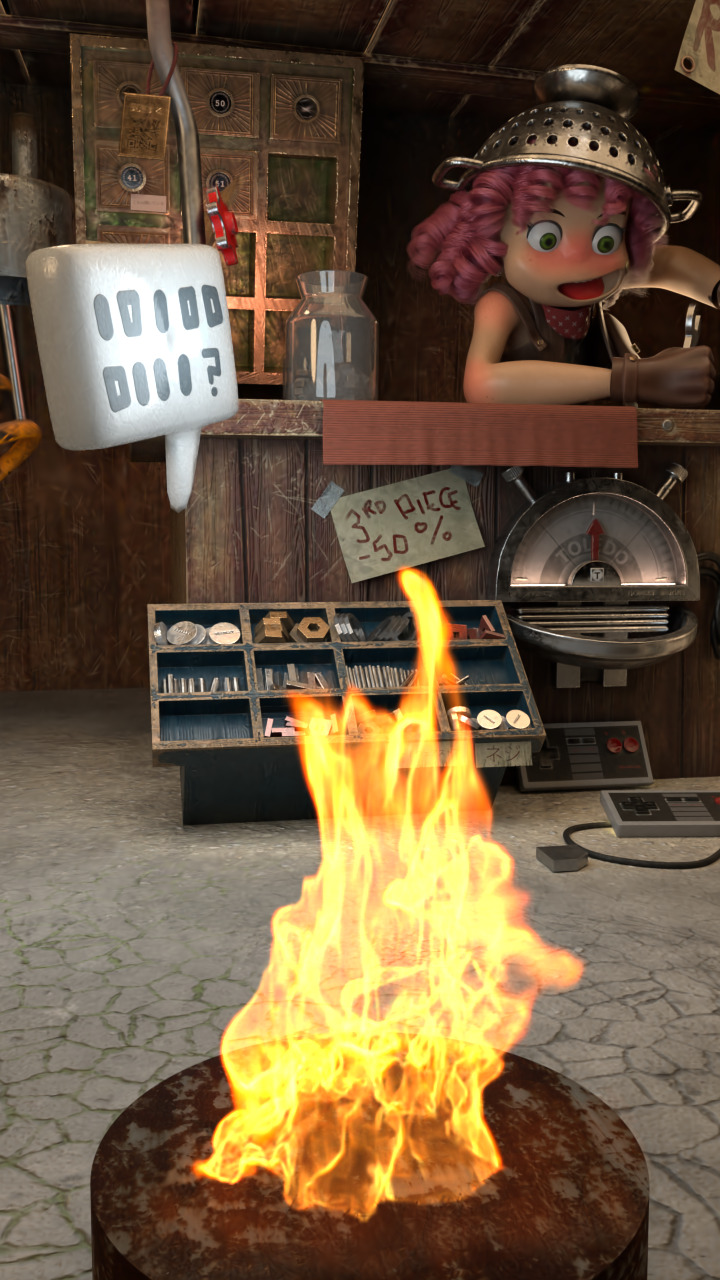}{\ang{0}}\hfill
  \rotpanel{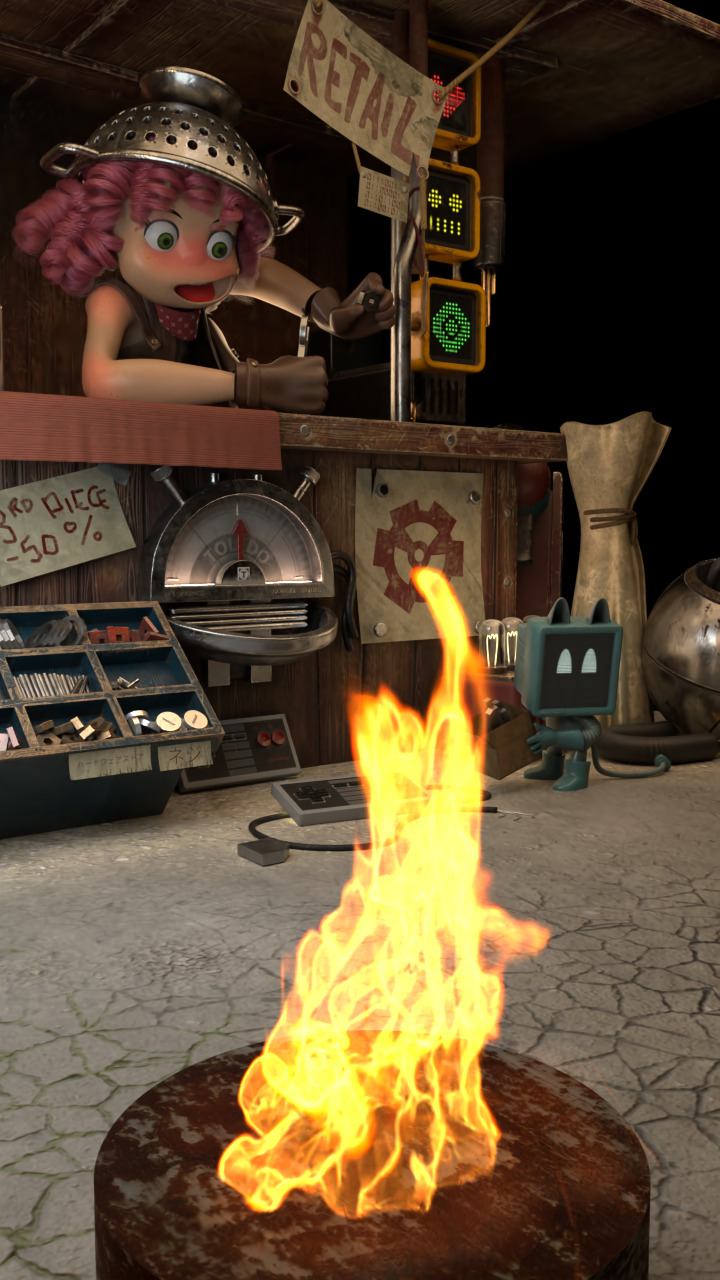}{\ang{25}}\hfill
  \rotpanel{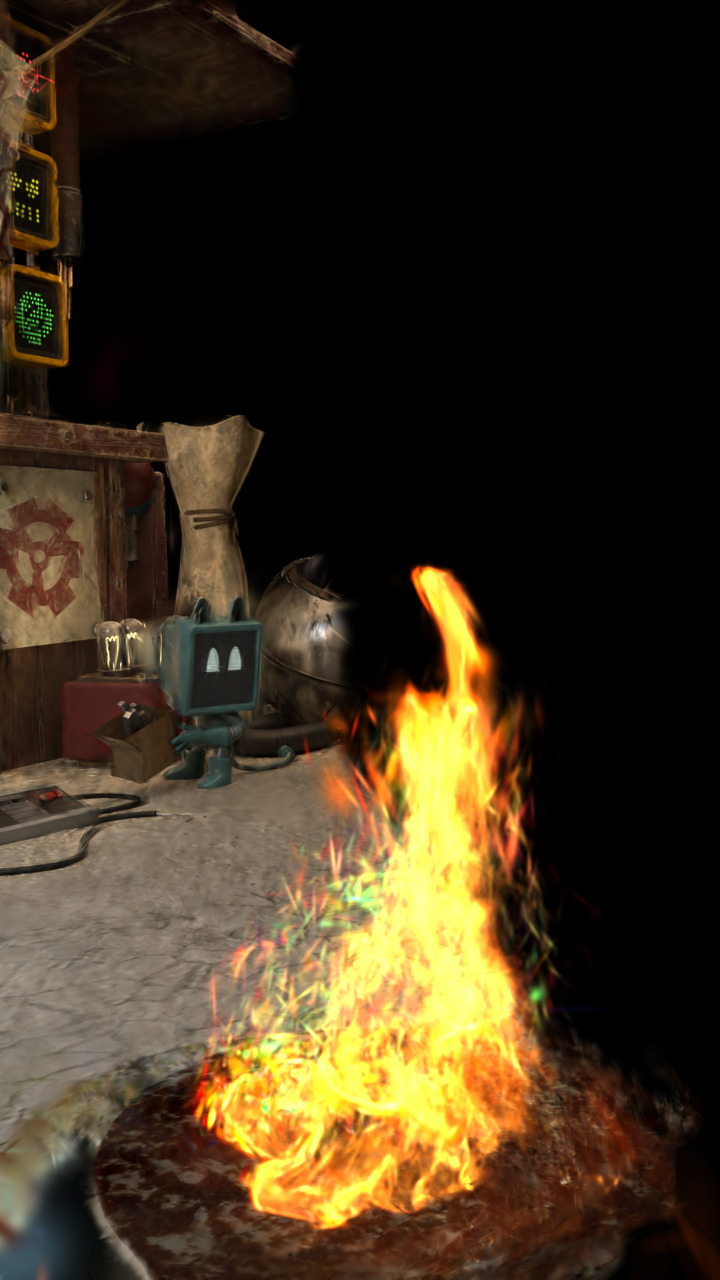}{\ang{50}}\hfill
  \rotpanel{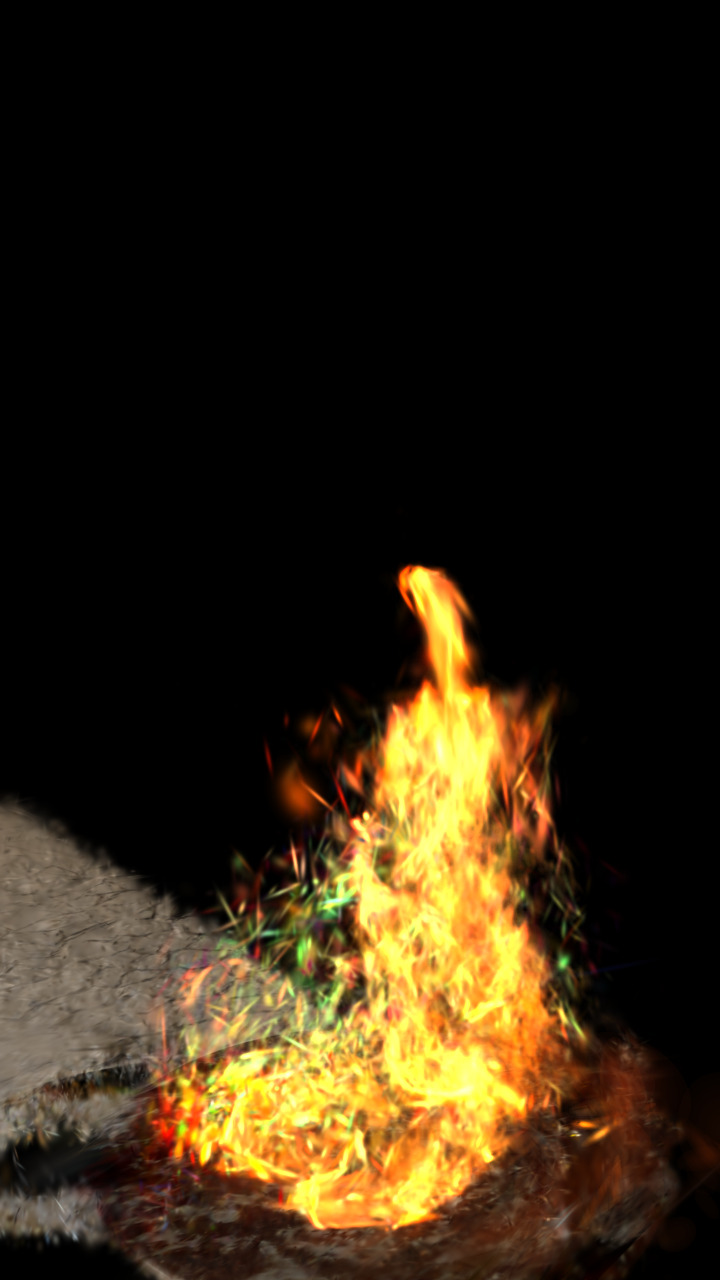}{\ang{75}}\hfill
  \rotpanel{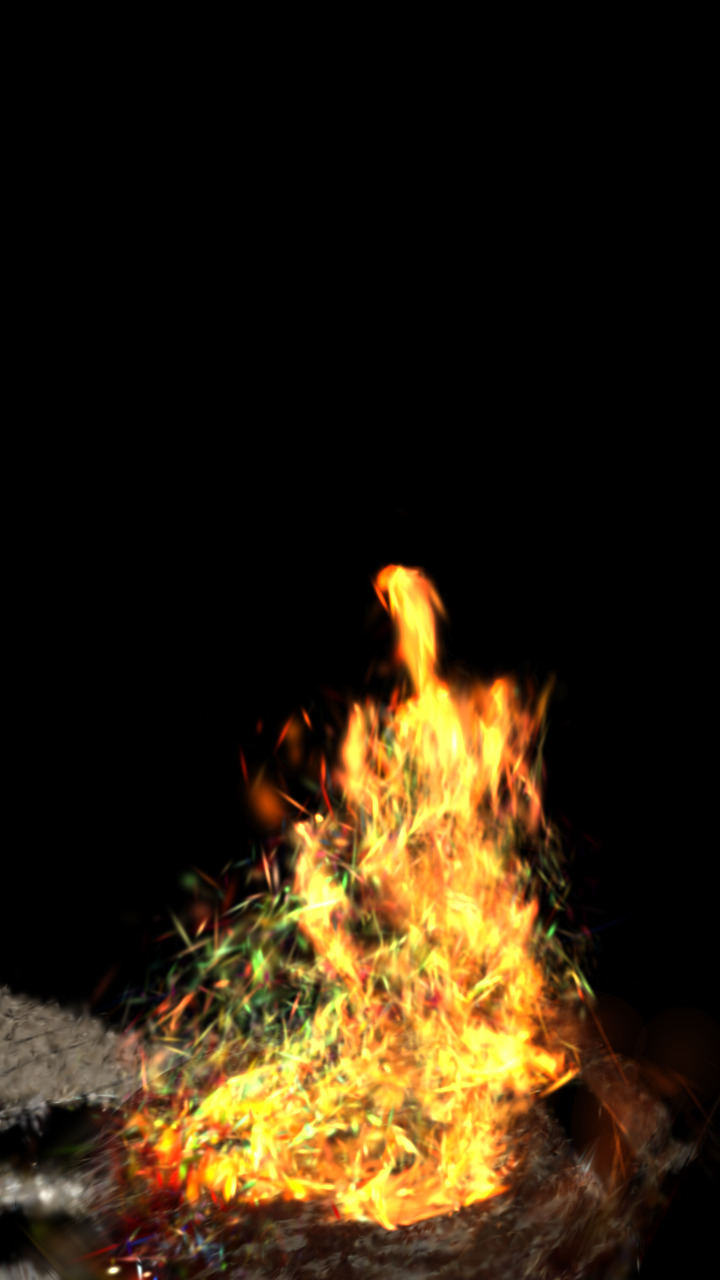}{\ang{100}}\hfill
  \rotpanel{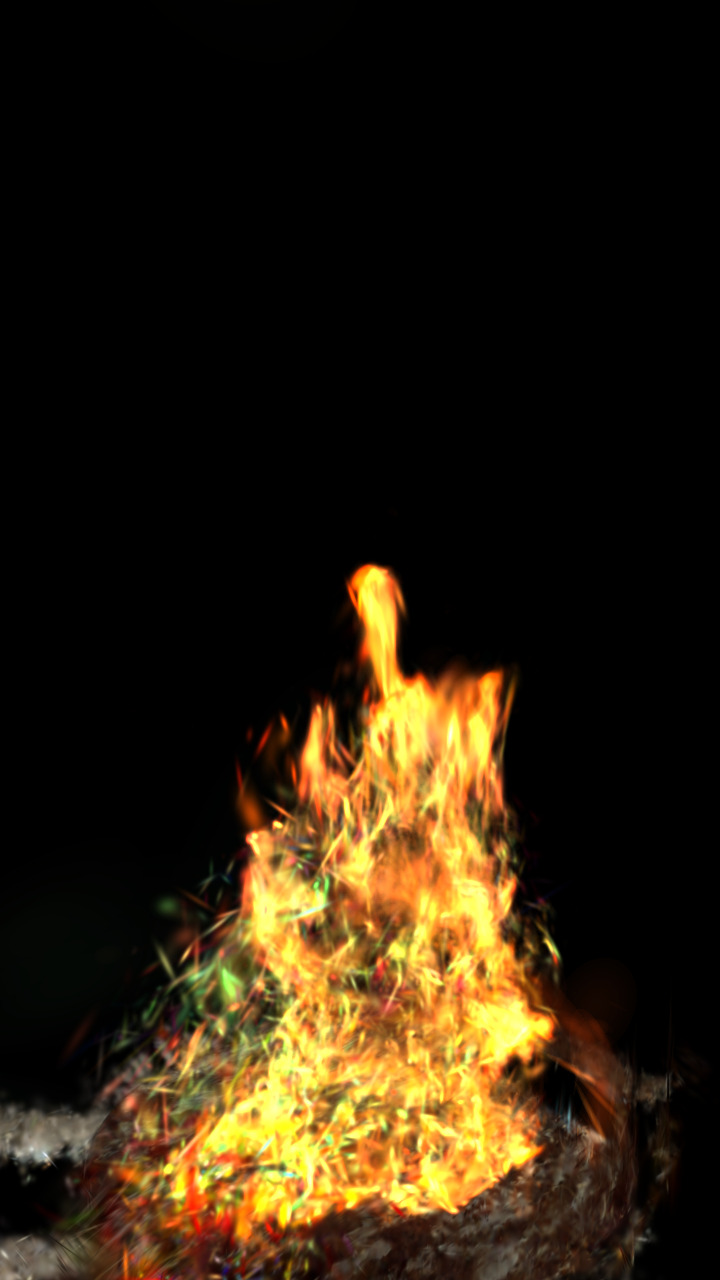}{\ang{125}}\hfill
  \rotpanel{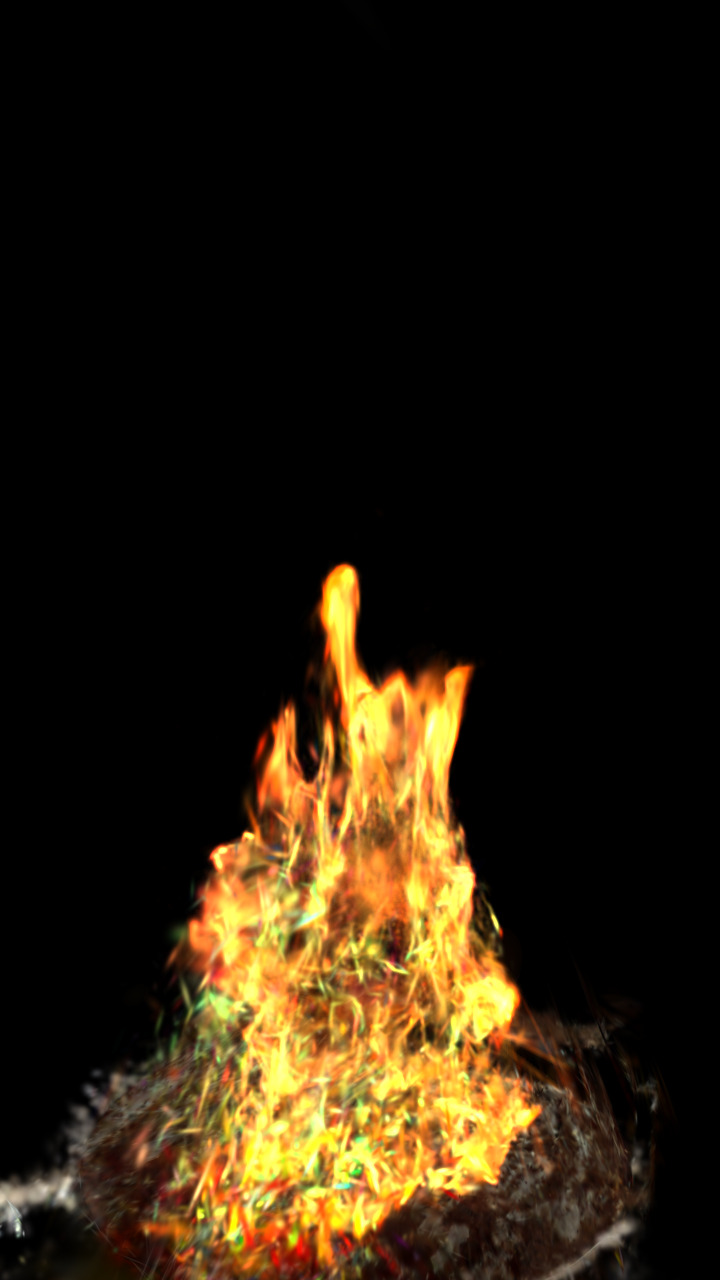}{\ang{150}}\hfill
  \rotpanel{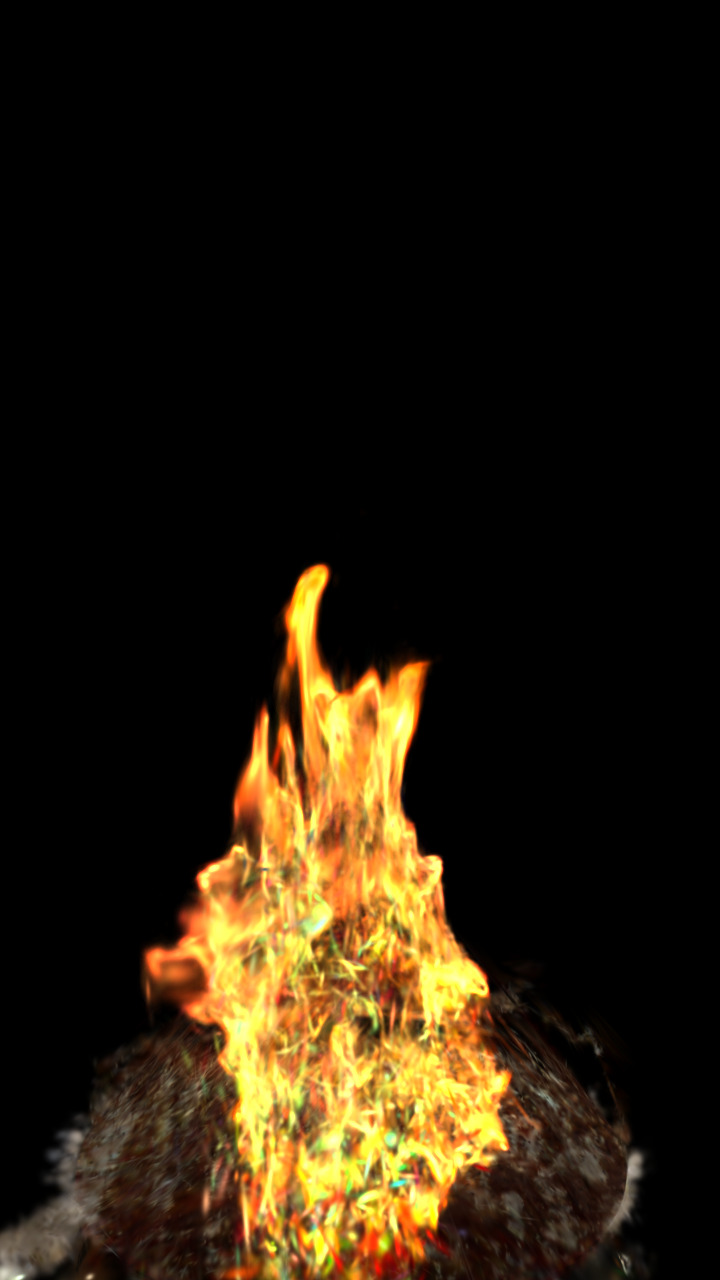}{\ang{175}}
  \caption{Reconstruction of the synthetic fire scene for highly novel views. The camera rotates around the scene, leaving the proximity of the training views. While this unmasks some artifacts in the less-supervised regions, the overall reconstruction remains visually and volumetrically plausible. There is no visible 3D to 2D billboard collapse in the flames.}
  \label{fig:rotation_strip}
\end{figure}

It is further necessary to demonstrate the graceful degradation of the reconstruction quality when completely leaving the proximity of the training views and entering highly novel views. To show this, we rendered a sequence of views along a circular trajectory around the fire, showing it from the side and from behind (see \cref{fig:rotation_strip}). As expected, this unmasks some artifacts in less-supervised regions of the scene, especially in Gaussians that are nearly fully occluded in the training views. However, the results clearly show that the reconstruction is not collapsing to a 2D billboard, but rather degrades gracefully while retaining the overall visual and volumetric plausibility of the scene. 

\section{Pipeline Visualization}
\label{supp:pipeline-visualization}
In our video, we present a visualization of our proposed reconstruction pipeline. For one scene, this video shows the relevant preprocessing and optimization steps. For each stage, we show the corresponding view for each of the three cameras. We start with the captured input videos and proceed with the process of removing the flames to obtain a static scene. Based on these images, we continue with the estimated depth maps and the Gaussian reconstruction of the static scene. For the dynamic scene, we first show the dense optical flow, followed by the optimization of the dynamic fire scene as the final stage.

\section{Implementation Details}
\label{supp:implementation-details}
For the static scene optimization, we applied the original 3DGS implementation~\cite{Kerbl2023GaussianSplatting}. To account for the underconstrained scene geometry, we increased the depth loss by a factor of \(100\). Furthermore, to avoid the collapse of the scene geometry due to limited view constraints, we disabled the pruning during the densification process. We trained for \(10^4\) iterations with standard optimization parameters.

For the dynamic optimization, we modified the original 3DGS implementation~\cite{Kerbl2023GaussianSplatting} by adding velocity and lifetime parameters, similar to FreeTimeGS~\cite{Wang2025FreeTimeGS}. Each Gaussian additionally received a boolean parameter indicating whether it is to be considered a static background Gaussian or a dynamic fire Gaussian.
During rendering, the temporal properties regarding position and opacity were purely applied to the Gaussians marked as dynamic. The optimization process including gradient-based optimization and densification was left as in the original implementation. 
To ensure a detailed reconstruction of the dynamic flames, we artificially limited the scale of the dynamic Gaussians by replacing their exponential with a sigmoid activation function, scaled by the voxel size used to initialize the scene.
To avoid dynamic Gaussians representing parts of the static background that undergo slight temporal variations, we apply an additional \textit{alpha loss} with a weight of \(0.1\). For this, we render the scene twice, first as the regular RGB scene, and secondly only the opacity of the dynamic Gaussians. Based on the rendered opacity and the pre-processed motion mask, we then compute the additional loss term.
For a strong initialization of the motion field, we reduced the spatial learning rate by a factor of \(10^{-2}\) for the trainable velocity parameter. We trained for \(3 \cdot 10^4\) iterations with a batch size of \(1\), evaluating on the test set at iterations \(1000\), \(7000\), and \(30000\). For the further evaluation and comparison, we utilized the trained parameters at the final iteration.

Regarding the baseline methods~\cite{Wu2024FourDGS,yang2023gs4d,10.5555/3737916.3741850}, we relied on the unmodified implementations provided by the authors. Adhering to the official implementation, we initialized all methods on COLMAP points and trained with the optimization parameters provided by the authors for the \emph{flame salmon} scene.
As there is no official implementation for FreeTimeGS~\cite{Wang2025FreeTimeGS}, we applied an unofficial implementation published on GitHub as \texttt{OpsiClear/FreeTimeGsVanilla}. Instead of initializing from COLMAP points, it computes pairwise RoMa~\cite{edstedt2024roma} matches for each pair of simultaneous frames and initializes the motion using k-nearest-neighbors on temporally adjacent per-frame point clouds. \cref{fig:roma} shows the issues with this initialization approach. While RoMa successfully predicts a dense model of the static background scene, large sections of the fire are not densely matched or even incorrectly matched to parts of the static scene.

\section{Code and Data}
\label{supp:code}
For code and data, we refer to the paper's official repository at \url{https://github.com/jna-358/gaussians_on_fire}.

\section{Compute Hardware}
\label{supp:compute-hardware}
For training, we used a system with a 16-core AMD Ryzen Threadripper PRO 3955, an NVIDIA RTX A6000 with \qty{48}{\giga\byte} of GPU memory, as well as \qty{128}{\giga\byte} of system memory. With this setup, training a single scene takes approximately \qty{30}{\minute} at a GPU memory usage of \qty{8}{\giga\byte} and a system memory usage of \qty{2}{\giga\byte}. Rendering a single image has been measured to take approximately \qty{7.4}{\milli\second}, resulting in a frame rate of \qty{131}{\hertz}.

\end{document}